\documentclass[10pt,twocolumn,letterpaper]{article}

\usepackage{cvpr}
\usepackage{times}
\usepackage{epsfig}
\usepackage{graphicx}
\usepackage{amsmath}
\usepackage{amssymb}

\usepackage{multirow}
\usepackage{xcolor}
\usepackage{enumitem}
\usepackage{colortbl}
\usepackage{booktabs}
\usepackage{mwe}
\usepackage{bm}
\newcommand\myparagraph[1]{\textbf{#1} ---}

\graphicspath{ {./images/} }

\def\eg{\emph{e.g}}

\def\etc{\emph{etc}}

\def\toolUrl{\url{https://reasoningpatterns.github.io}}

\definecolor{colorh}{rgb}{0.8, 0.8, 0.8}
\definecolor{colorreason}{rgb}{0.30, 0.45, 0.69}
\definecolor{colorbias}{rgb}{0.87, 0.52, 0.32}
\definecolor{colorother}{rgb}{0.34, 0.66, 0.41}

\newcommand{\qemph}[1]{``\emph{#1}''}

\newcommand\mydots{\makebox[0.7em][c]{.\hfil.\hfil.}}

\newcommand{\cam}[1]{{#1}}

\usepackage[pagebackref=true,breaklinks=true,letterpaper=true,colorlinks,bookmarks=false]{hyperref}

\cvprfinalcopy % *** Uncomment this line for the final submission

\def\cvprPaperID{990} % *** Enter the CVPR Paper ID here

\ifcvprfinal\fi
\begin{document}

%%%%%%%%% TITLE
\title{How Transferable are Reasoning Patterns in VQA?}

\author{
	Corentin~Kervadec$^{1,2}$\thanks{Both authors contributed equally.} \quad Th\'eo Jaunet$^{2*}$ \quad Grigory~Antipov$^1$\\
	Moez~Baccouche$^1$ \quad Romain Vuillemot$^{2}$ \quad Christian~Wolf$^2$ \\
	$^1$Orange, Cesson-S\'evign\'e, France \quad $^2$LIRIS, INSA - École Centrale, Lyon, UMR CNRS 5205, France\\
	{\tt\small corentinkervadec.github.io theo-jaunet.github.io christian.wolf@insa-lyon.fr}\\
	{\tt\small \{grigory.antipov, moez.baccouche\}@orange.com romain.vuillemot@gmail.com}
}

\maketitle

%%%%%%%%% ABSTRACT
\begin{abstract}
\noindent
Since its inception, Visual Question Answering (VQA) is notoriously known as a task, where models are prone to exploit biases in datasets to find shortcuts instead of performing high-level reasoning. Classical methods address this by removing biases from training data, or adding branches to models to detect and remove biases. In this paper, we argue that uncertainty in vision is a dominating factor preventing the successful learning of reasoning in vision and language problems. We train a visual oracle and in a large scale study provide experimental evidence that it is much less prone to exploiting spurious dataset biases compared to standard models. We propose to study the attention mechanisms at work in the visual oracle and compare them with a SOTA Transformer-based model. We provide an in-depth analysis and visualizations of reasoning patterns obtained with an online visualization tool which we make publicly available\footnote{\toolUrl}. 
{We exploit these insights by transferring reasoning patterns from the oracle to a SOTA Transformer-based VQA model taking standard noisy visual inputs via fine-tuning.}
In experiments we report  higher overall accuracy, as well as accuracy on infrequent answers for each question type, which provides evidence for improved generalization and a decrease of the dependency on dataset biases.
\end{abstract}

%%%%%%%%% BODY TEXT
\section{Introduction}

\begin{figure}[t]
\centering
\includegraphics[width=\linewidth]{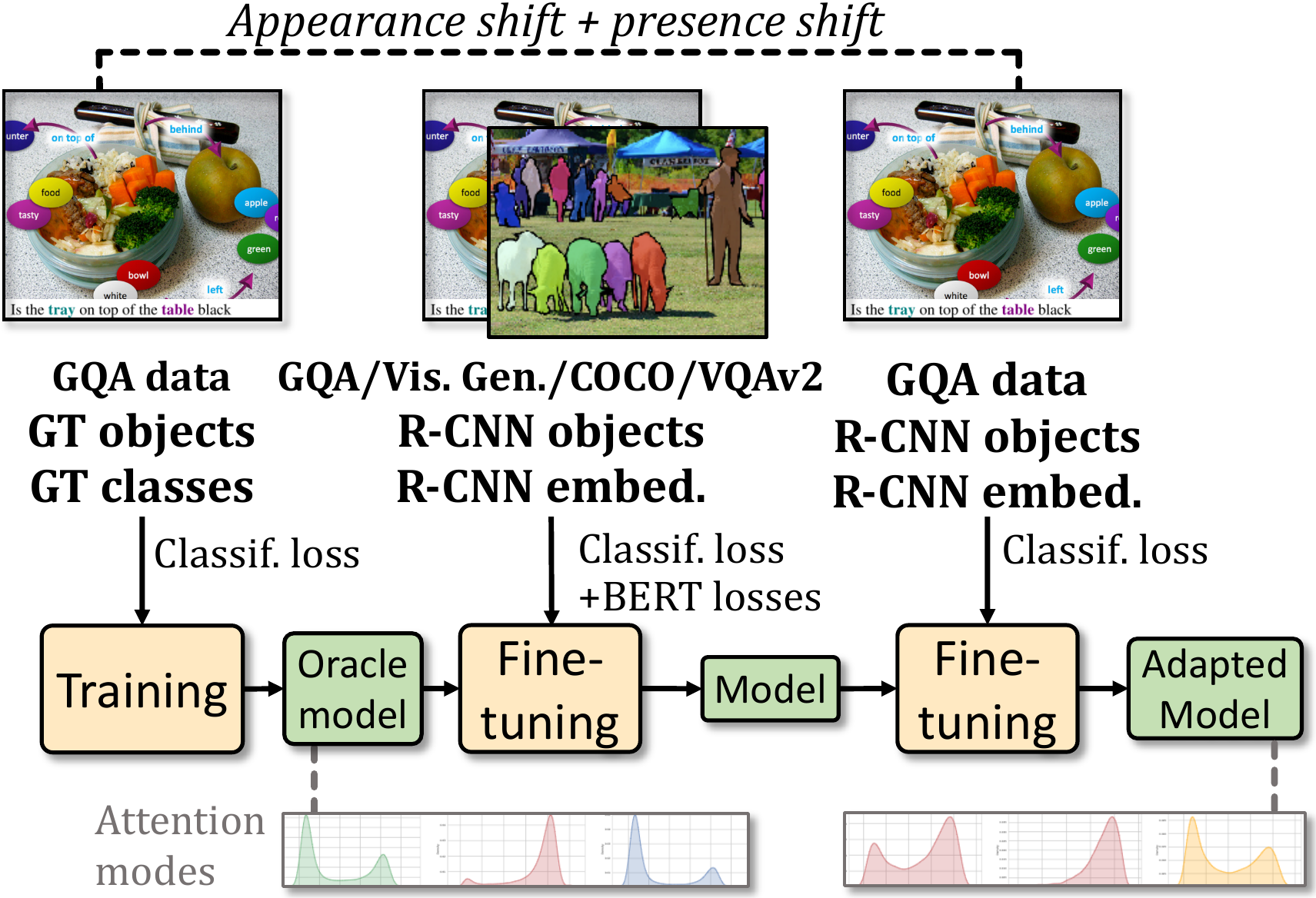}
\caption{\label{fig:teaser}We argue that noise and uncertainties in visual inputs are the main bottleneck in VQA preventing successful learning of reasoning capacities. In a deep analysis, we show that oracle models with perfect sight, trained on noiseless visual data, tend to depend significantly less on bias exploitation. We exploit this by training models on data without visual noise, and then transfer the learned reasoning patterns to real data. We illustrate successful transfer by an analysis and visualization of attention modes.}
\end{figure}

\noindent
The high prediction performance obtained by high-capacity deep networks trained on large-scale data has led to questions concerning the nature of these improvements. Visual Question Answering (VQA) in particular has become a testbed for the evaluation of the reasoning and generalization capabilities of trained models, as it combines multiple modalities of heterogeneous nature (images and language) with open questions and large varieties.
It has been shown, that current models are prone to exploiting harmful biases in the data, which can provide unwanted shortcuts to learning in the form of ``\emph{Clever Hans}'' effects \cite{teney2020value,RosesAreRed}.

In this work we study the capabilities of VQA models to \qemph{reason}. An exact definition of this term is difficult, we refer to \cite{bottou2014machine,RosesAreRed} and define it as \qemph{algebraically manipulating words and visual objects to answer a new question}. In particular, we interpret reasoning as the opposite of exploiting spurious biases in training data. We argue, and in Section~\ref{sec:vis_attention} will provide evidence for this, that learning to algebraically manipulate words and objects is difficult when visual input is noisy and uncertain compared to learning from perfect information about a scene.
When objects are frequently missing, detected multiple times or recognized with ambiguous visual embeddings wrongly overlapping with different categories, relying on statistical shortcuts may be an easy shortcut for the optimizer
\footnote{See \cite{geirhos2020shortcut} for an interesting review of \emph{shortcut learning}.}
We show, that a perfect-sighted oracle model learns to predict answers while significantly less relying on biases in training data.
We claim that once any noise has been removed from visual input, replacing object detection output by Ground Truth (GT) object annotations, a deep neural network can more easily learn the reasoning patterns required for prediction and for generalization.

In the line of recent work in AI explainability \cite{Lipton2016TheInterpretability,ribeiro2016should}, and data visualization \cite{hohman2018visual,vig2019transformervis,derose2020attention}, we propose an in-depth analysis of attention mechanisms in Transformer-based models and provide indications of the patterns of reasoning employed by models of different strengths. We visualize different operating modes of attention and link them to different sub tasks (\qemph{functions}) required for solving VQA. In particular, we use this analysis for a comparison between oracle models and standard models processing noisy and uncertain visual input, highlighting the presence of reasoning patterns in the former and less so in the latter.

Drawing conclusions from this analysis, we propose to 
{fine-tune the perfectly-sighted oracle model on the real noisy visual input (see Fig. \ref{fig:teaser}).}
Using the same analysis and visualization techniques, we show that attention modes absent from noisy models are transferred successfully from oracle models to deployable
\footnote{{\emph{Deployable}: the model \textbf{does not} use ground-truth visual inputs.}}
models, and we report improvements in overall accuracy and generalization.

\myparagraph{Contributions}
(i) An in-depth analysis of reasoning patterns at work in Transformer-based models, comparing Oracles vs. deployable models, including visualization of attention modes; an analysis of the relationships between attention modes and reasoning, and the impact of attention pruning on reasoning.
{(ii) We propose to transfer reasoning capabilities, learned by the oracle, to SOTA VQA methods with noisy input and improve overall performance and generalization on the GQA~\cite{hudson2019gqa} dataset}
(iii) We show that this transfer is complementary with self-supervised large-scale pre-training (LXMERT~\cite{tan2019lxmert}/BERT-like).

% -------------------------------------------------------------------------

\section{Related work}
\myparagraph{Visual Question Answering (VQA)}
as a task was introduced in various datasets, such as VQAv1~\cite{antol2015vqa} and VQAv2~\cite{goyal2017making} (built from human annotators), or CLEVR~\cite{johnson2017clevr} and GQA~\cite{hudson2019gqa} (automatically-generated from fully-synthetic and real-world images, respectively).
Additional splits were proposed to evaluate specific reasoning capabilities.
For instance, VQA-CP~\cite{vqa-cp} explicitly inverts the answer distribution between train and test splits. 
Following recent critics and controversies about these evaluations~\cite{teney2020value, shrestha2020negative}, the GQA-OOD dataset~\cite{RosesAreRed} introduced a new split of GQA focusing on rare (Out-Of-Distribution / OOD) question-answer pairs, and showed that many VQA models strongly rely on dataset biases.
This growing amount of diverse datasets has been accompanied by the development of more sophisticated VQA models.
While an exhaustive survey of methods is out of the scope of this paper, 
one can mention families based on object-level attention~\cite{anderson2018bottom}, bilinear fusion~\cite{kim2018bilinear}, tensor decomposition~\cite{ ben2017mutan}, neural-symbolic reasoning~\cite{yi2018neural}, neural~\cite{andreas2016neural} and  meta~\cite{chen2021meta} module networks. 

\myparagraph{Transformers and Vision-Language reasoning}
In this work, we focus on Transformers~\cite{vaswani2017attention} due to their wide adoption and their powerful attention mechanism.
MCAN~\cite{yu2019deep} and DFAF~\cite{gao2019dynamic}  introduced the use of object-level self-attention and co-attention mechanisms to model intra- and inter-modality interactions in VQA.
More recent work~\cite{kervadecweak,chen2020uniter,lu2019vilbert,tan2019lxmert} suggests that the combination of Transformers with a large-scale BERT~\cite{devlin2019bert}-like pretraining can be beneficial for VQA.
Self-attention on pixel-level \cite{16x16ICLR2021,SSTVOSCVPR2021} is, but up to our knowledge, not used done for VQA.

\myparagraph{Attention and reasoning patterns in Transformers}
Analysis of self-attention mechanisms has received considerable attention recently.
In~\cite{vig2019multiscale},  a visualization tool for analysing BERT attention layers is introduced.
\cite{derose2020attention} studies how training strategies and fine-tuning impact  attention in BERT-like models.
Voita et al.~\cite{voita2019analyzing} classifies BERT's attention heads according to their functionality, reporting  a significant simplification of the model's complexity via pruning.
Finally, \cite[Appendix A.5]{ramsauer2020hopfield}
measures energy distributions and classifies them based on their meta-stable states.

Following this work in NLP, similar studies have appeared in vision and language.
\cite{kervadecweak} explores the emergence of word-object alignment in the attention maps when adding a weakly supervised objective. \cite{cao2020behind, li2020does} study to what extent the attention maps in BERT-like pretrained VQA Transfomers encode various vision-language information.
While these methods provide a better understanding of the amount of information captured by VQA models, they do not shed light on how this information is used. In our work, we analyze how various VQA tasks are encoded in different attention heads. To this end, we apply an energy-based analysis inspired by~\cite{ramsauer2020hopfield}. In addition, we study attention in perfect-sighted oracle Transformers in order to identify which patterns lead to better reasoning.
Our findings lead to an \emph{Oracle Transfer} strategy, which allows to improve performance and generalization in standard transformer  models.
Finally, our work is related to \cite{dumoulin2018feature-wise}, which
found evidence for relationships between questions and the modulation of (non transformer) model parameters on the synthetic CLEVR~\cite{johnson2017clevr} dataset.

% -------------------------------------------------------------------------
\section{Analysis of Reasoning Patterns}
\label{sec:vis_attention}

\noindent
In this section we will analyze reasoning behavior in Transformer-based VQA models and make the case for the impact of training on visual GT data. First, to make the paper self-contained, we provide a short introduction into Vision-Language (VL)-Transformers as proposed in standard literature~\cite{yu2019deep, gao2019dynamic, tan2019lxmert,lu2019vilbert, li2019visualbert, chen2020uniter, su2020vl}.

Given two different modalities, Vision (V) and Language (L), VL-Transformers are composed of the succession of intra-modality $T_{-}^{V}(\cdot)$, $T_{-}^{L}(\cdot)$ and inter-modality $T_{\times}^{V\leftarrow L}(\cdot,\cdot)$, $T_{\times}^{L\leftarrow V}(\cdot,\cdot)$ multi-head attention layers~\cite{vaswani2017attention}.
As defined in the seminal paper~\cite{vaswani2017attention}, multi-head attention layers $T(\cdot)$ (both intra- and inter-modality ones) can be expressed as a set of self-attention layers $t(\cdot)$ which are performed in parallel on several ``heads''.
For example, given an input sequence $\bm{v}$ of the visual embeddings, a visual intra-modality $n$-head attention layer $T_{-}^{V}(\cdot)$ performs as a set of $h$ visual intra-modality self-attention layers $\{{t_{-}^{V}}^{(1)}(\cdot),\mydots,{t_{-}^{V}}^{(h)}(\cdot)\}$, the outputs of which are concatenated and then combined:
\begin{equation}
    T_{-}^{V}(\bm{v})=
    \left [ 
    \ {t_{-}^{V}}^{(1)}(\bm{v}), \ \mydots, \ {t_{-}^{V}}^{(h)}(\bm{v}) \
    \right ]
    W^{\bm{O}}
\end{equation}
where $W^{\bm{O}}$ is a trainable matrix which is particular for each multi-head attention layer.
Each layer $t_{-}(\cdot)$ is defined on a set of input (vision or language) embeddings $\bm{x}$ of the same dimension $d$ as
\begin{equation}
    t_{-}(\bm{x})=\sum_{j}\bm{\alpha}_{ij}\bm{x}^v_{j},
\end{equation}
where the query $\bm{x}^q$, key $\bm{x}^k$ and value $\bm{x}^v$ matrices are given as follows: $\bm{x}^q{=}\bm{W}^{q}\bm{x}$, $\bm{x}^k{=}\bm{W}^{k}\bm{x}$ and $\bm{x}^v{=}\bm{W}^{v}\bm{x}$. 
All $\bm{W}^{.}$ are trainable parameters.
In particular, $\bm{x}^q$ and $\bm{x}^k$ are used to calculate the self-attention weights $\bm{\alpha}_{\cdot j}$ as follows:
\begin{equation}
\bm{\alpha}_{\cdot j}{=}(\bm{\alpha}_{1j},\mydots,\bm{\alpha}_{ij},\mydots,\bm{\alpha}_{nj}){=}\sigma(\mydots,\frac{{\bm{x}^{q}_{i}}^{T}\bm{x}^{k}_{j}}{\sqrt{d}},\mydots),
\end{equation}
with $\sigma$ being the softmax operator.
In this paper, we mainly focus on the attention maps $\{\alpha_{ij}\}$ which are composed of these self-attention weights.

Finally, inter-modality self-attention layers $t_{\times}(\cdot,\cdot)$ are defined in the same way, as the intra-modality ones, but unlike the latter they calculate queries, keys and values on two sets of input embeddings of different modalities.
More precisely, for the self-attention layer $t_{\times}^{V\leftarrow L}(\bm{v},\bm{l})$, the query matrix $\bm{v}^q=\bm{W}^{q}\bm{v}$ is calculated on vision embeddings, while the key $\bm{l}^k=\bm{W}^{k}\bm{l}$ and the value matrices $\bm{l}^v=\bm{W}^{v}\bm{l}$ are calculated on the language ones.
For the $t_{\times}^{L\leftarrow V}(\bm{l},\bm{v})$ self-attention layer, the matrices are calculated symmetrically ($\bm{l}^q=\bm{W}^{q}\bm{l}$, $\bm{v}^k=\bm{W}^{k}\bm{v}$ and $\bm{v}^v=\bm{W}^{v}\bm{v}$, respectively). In addition, each $t_{\times}^{V\leftarrow L}$ (resp. $t_{\times}^{L\leftarrow V}$) is followed by a self-attention $t_{-}^{V}$ (resp. $t_{-}^{L}$).
A general view of the architecture is available in the supp. mat..

\myparagraph{Experimental setup}
All analyses in this section have been performed with a hidden embedding size $d=128$ and a number of per-layer heads $h=4$. This corresponds to a tiny version of the architecture used in LXMERT~\cite{tan2019lxmert} where $d=768$ and $h=12$.
Therefore, ``tiny-LXMERT'' corresponds to the VL-Transformer architecture plus  BERT-like (LXMERT) pre-training.
Unless specified otherwise, objects have been detected with Faster R-CNN~\cite{ren2015faster}.
{Visualizations are done on GQA~\cite{hudson2019gqa} (validation set) as
it is particularly well suited for evaluating a large variety of reasoning skills. However, as GQA contains synthetic questions constructed from pre-defined templates, the dataset only offers a constrained VQA environment. Additional experiments might be required to extend our conclusions to more natural setups.}

\subsection{Visual noise vs. models with perfect-sight}

\noindent
We conjecture that difficulties in the computer vision pipeline 
are the main cause preventing VQA models in learning to reason well, and which leads them to exploit spurious biases in training data. Most of these methods use pre-trained off-the-shelf object detectors during training and evaluation steps. But in a significant number of cases, the visual objects necessary for reasoning are misclassified, or even not detected at all, as indicated by detection rates of SOTA detectors on the Visual Genome dataset~\cite{krishna2017visual}, for instance.
Under these circumstances, even a perfect VQA model is unable to predict correct answers without relying on statistical shortcuts. 

To further explore this working hypothesis, we trained an oracle model with perfect sight, \emph{i.e} a model which receives perfect visual input, and compare it with tiny-LXMERT. Based on the same VL-Transformer, it receives the GT objects from the GQA annotations, encoded as GT bounding boxes and 1-in-K encoded object classes replacing the visual embeddings of the classical model. All GT objects are fed to the model, not only objects required for reasoning.
We study the capabilities of both models, the oracle model and the classical one, to \qemph{reason}. Following~\cite{RosesAreRed} we measure the reasoning capabilities of a VQA model as the capacity to correctly answer questions, where the GT answer is rare w.r.t. the question group, \emph{i.e} the type of questions being asked. We evaluate the models on the GQA-OOD benchmark~\cite{RosesAreRed} designed for OOD evaluation.

\begin{figure}[t] \centering
    \includegraphics[width=0.9\linewidth]{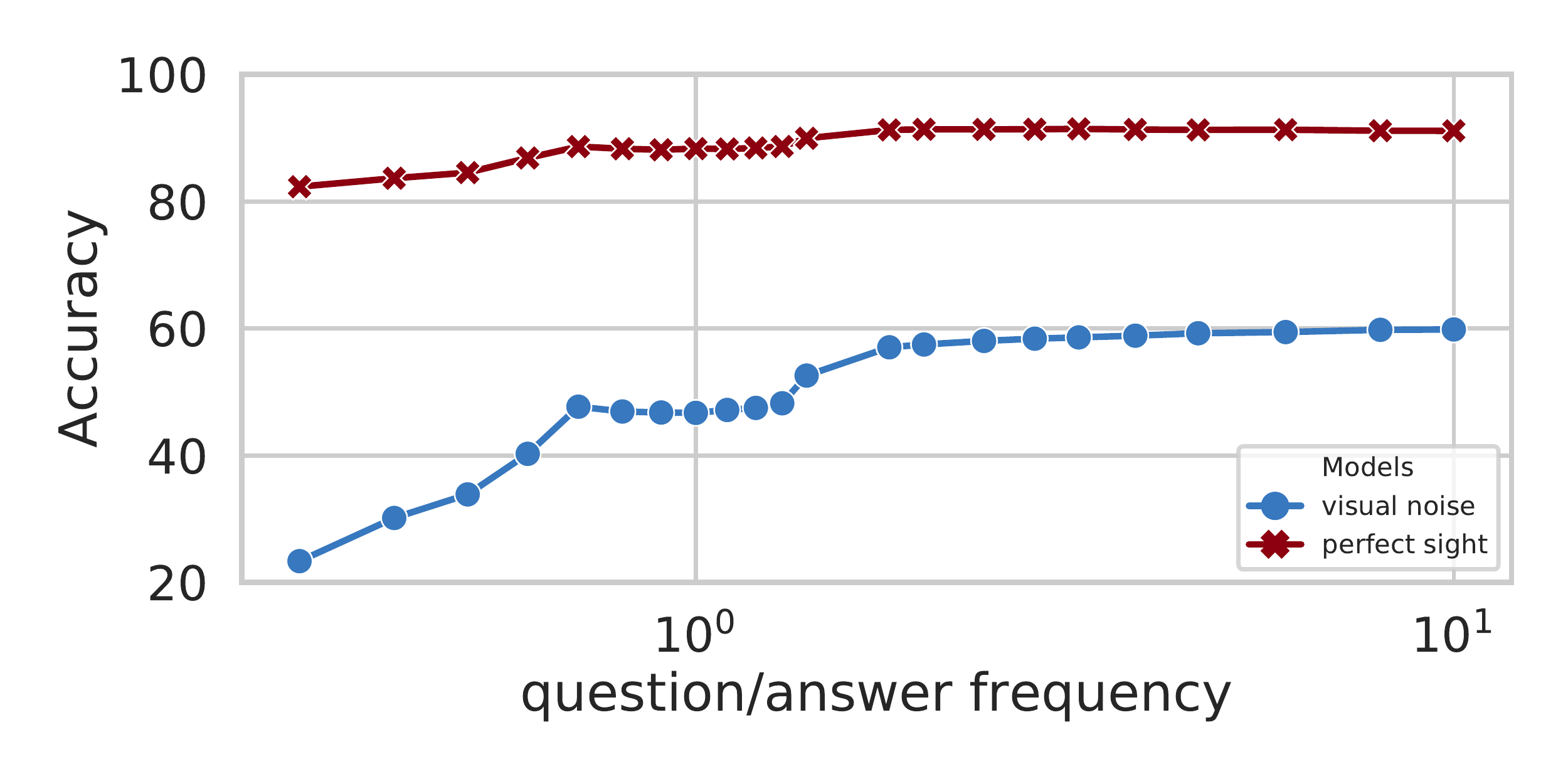}
    \caption{\label{fig:ood}Uncertainties and noise in visual input dominate the difficulties in learning reasoning: comparison of the out-of-distribution generalization between two different VQA Models. A perfectly-sighted oracle model and a standard noisy vision based model trained on the GQA-OOD benchmark~\cite{RosesAreRed}. For the classical model, accuracy drops for questions where the GT answer is rare (left side) compared to frequent answers (right side), indicating probable bias exploitation. In contrast, the oracle obtains high performance also on rare answers. Both models are ``tiny-LXMERT''.}
\end{figure}

Fig.~\ref{fig:ood} illustrates the model behavior in different situations. At the extreme case (left side of the plot), the model is evaluated on the rarest samples only, while on the right side all samples are considered. We observe that the performance of the classical model taking noisy visual (tiny-LXMERT) drops sharply for (image, question) pairs with rare GT answers, which is an indication for a strong dependency on dataset biases. We would like to insist that in this benchmark the rarity of a GT answer is determined w.r.t. the question type, which allows to measure biases taking into account language.
The oracle model, on the other hand, obtains performances which are far less dependent on answer rarity, providing evidence for its ability to overcome statistical biases. As a consequence, we conjecture that the visual oracle is closer to a real `reasoning process', by predicting answer resulting from a manipulation of words and objects, rather than by having captured statistical shortcuts. In the absence of GT on reasoning, we admit that there is no formal proof to this statement, but we believe that the evidence above is sufficient.

\begin{figure}[t] \centering
{\footnotesize
\begin{tabular}{ccc}
    \begin{minipage}{2.2cm}
        \includegraphics[width=\linewidth]{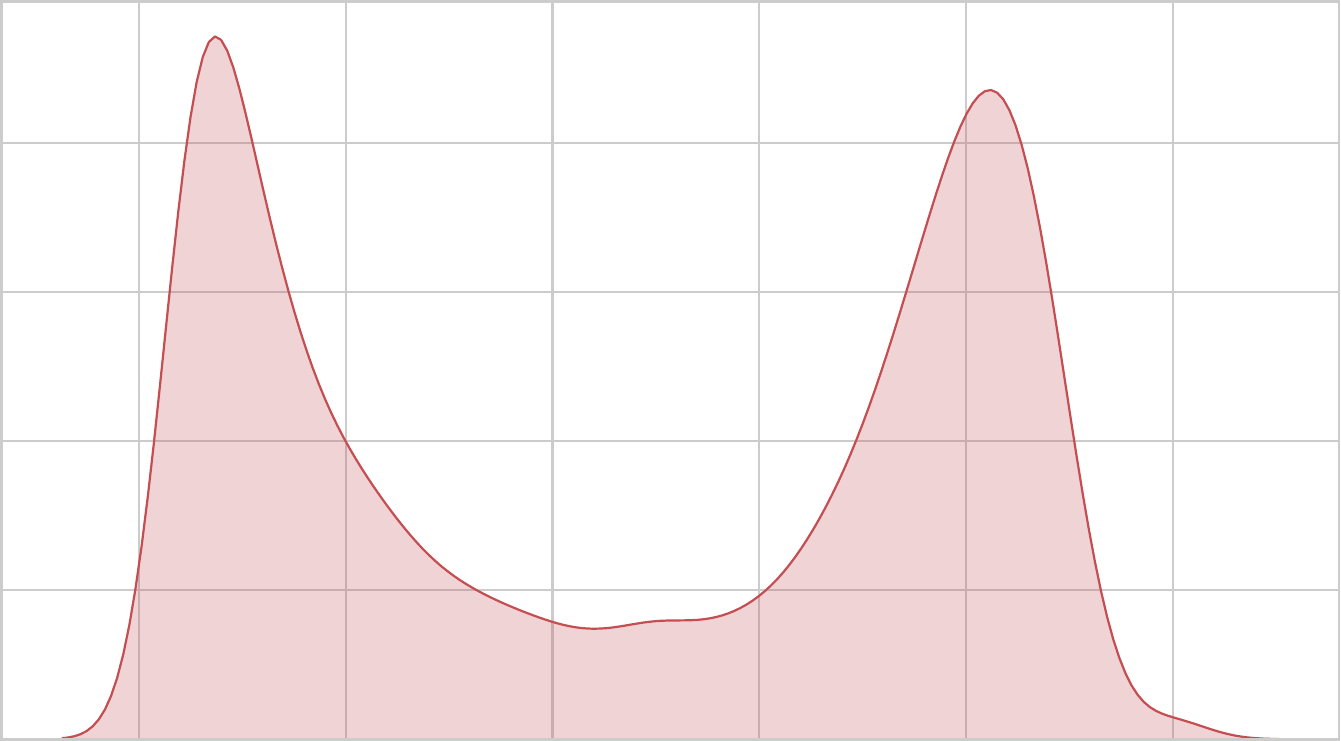}
    \end{minipage}
    &
    \begin{minipage}{2.2cm}
        \includegraphics[width=\linewidth]{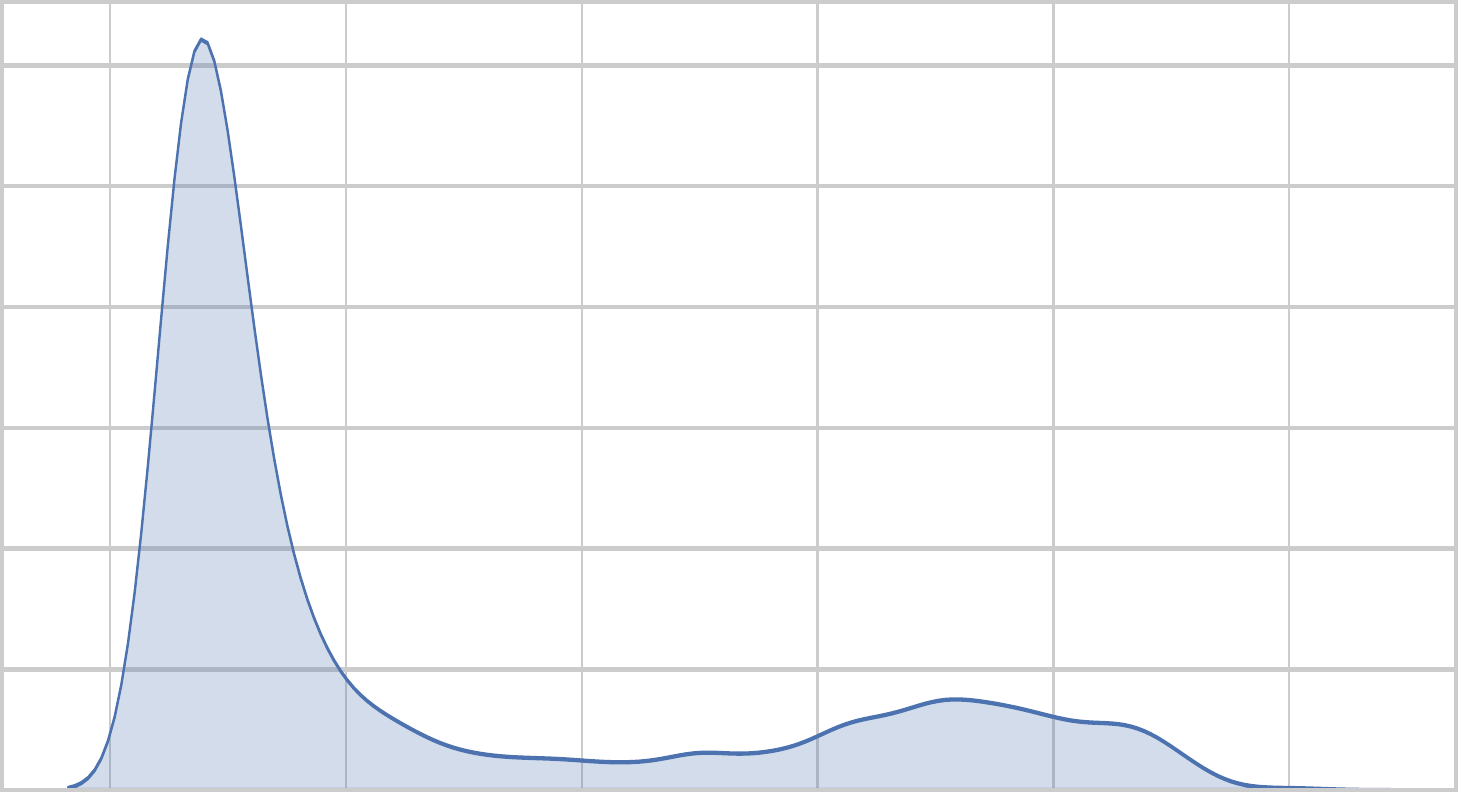}
    \end{minipage}
    &
    \begin{minipage}{2.2cm}
        \includegraphics[width=\linewidth]{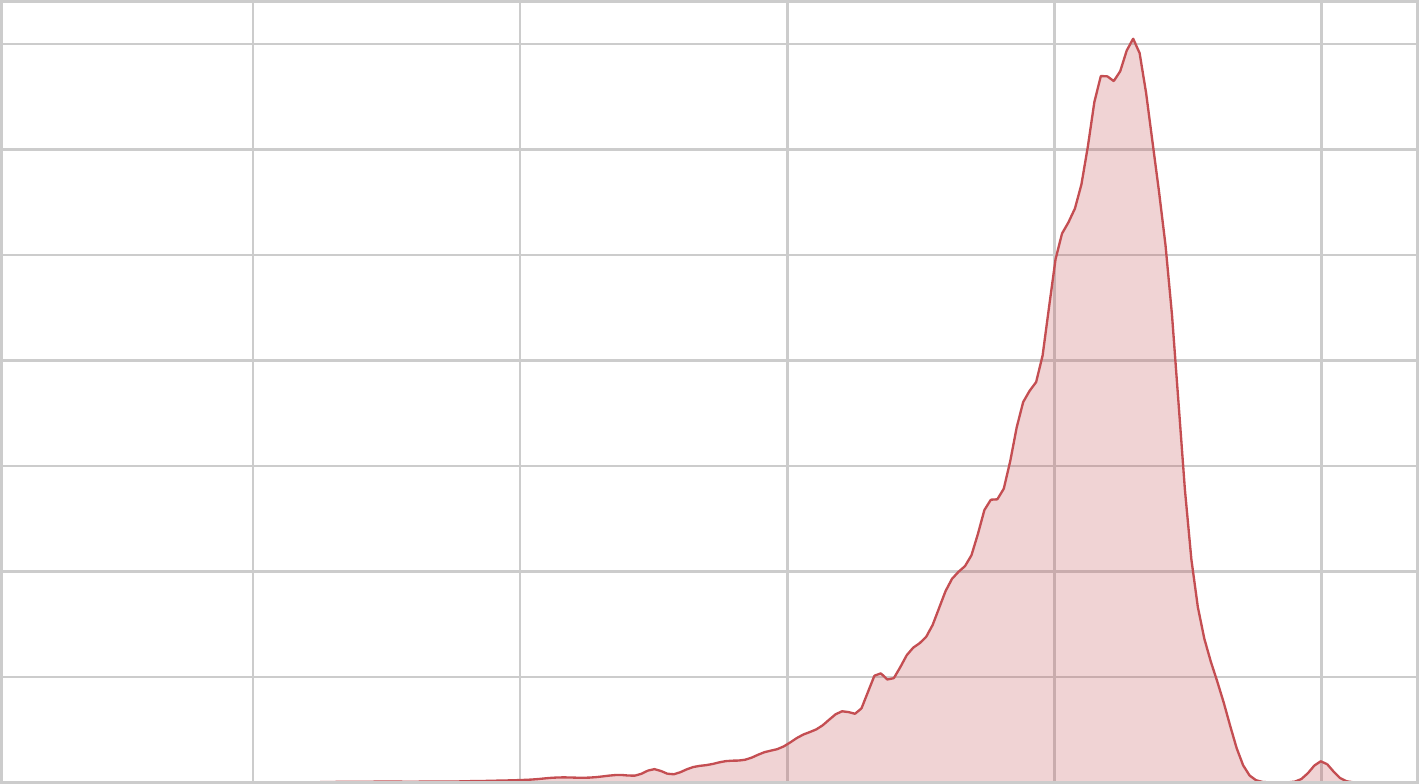}
    \end{minipage}
    \\
    (a) Bimorph & (b) Dirac & (c) Uniform
    \\
\end{tabular}
}
\caption{\label{fig:attention_modes}Attention modes learned by the oracle model. Following~\cite{ramsauer2020hopfield}, for each head we plot the distribution of the number $k$ of tokens required to reach $90\%$ of the attention energy (GQA-val). X-axis (from $0$ to $100\%$): ratio of the tokens $k$ w.r.t. the total number of tokens.
Plots are not attention distributions, but distributions of indicators of attention distributions. We observe three major modes: (a) \qemph{bimorph} attention, unveil two different types of attention distribution for the same head; (b) dirac attention with high $k$-median, \emph{i.e} small meta stable state; (c) uniform attention, with low $k$-median, \emph{i.e} very large meta stable state. }
\vspace*{-4mm}
\end{figure}

\subsection{Attention modes in VL-Transformers}
\label{sec:attention_modes}

\noindent
Attention distributions are at the heart of the VL-Transformer. They are not directly supervised during training, their behavior emerges from training the different VQA objectives, \emph{i.e} the discriminative loss as well as the eventual additional BERT-like objectives~\cite{tan2019lxmert}. Their definition as a strength of association between different items makes them a prime candidate for visualization of inner workings of deep models.
We analyze attention, and in particular we observe different attention modes in trained VQA models.

Following~\cite{ramsauer2020hopfield}
, we visualize the distribution of attention energy associated with each Transformer head in multi-headed attention. For each attention map, associated with a given head for a given sample, we calculate the number $k$ of tokens required to reach a total sum of $90\%$ of the distribution energy. A low $k$-number is caused by peaky attention, called \emph{small meta-stable state} in~\cite{ramsauer2020hopfield}, while a high $k$-number indicates uniform attention, close to an average operation (\emph{very large meta-stable state}).  For each head, and over a subset of validation samples, we plot the distribution of $k$-numbers, and for some experiments we summarize it with a median value taken over samples and over tokens.

\myparagraph{Diversity in attention modes}
In this experiment we focus on the oracle VL-Transformer, where we observed a high diversity in attention modes. 
We also observed that some layers' heads, especially those processing the visual modality  ($t_{-}^{V}$ or $t_{\times}^{V\leftarrow L}$) are mainly working with close-to-average attention distributions (\emph{very large meta-stable states}~\cite{ramsauer2020hopfield}). On the other hand, we observed smaller meta-stable states in the language layers ($t_{-}^{L}$ or $t_{\times}^{L\leftarrow V}$).
{This indicates that the reasoning process in the oracle VL-Transformer is in large part executed by the model as a transformation of the language features, which are successively contextualized (i.e. influenced) by the visual features (and not the opposite).}

\begin{figure}[t] \centering
{\footnotesize
\begin{tabular}{cc}
    \begin{minipage}{3.7cm}
        \includegraphics[width=\linewidth]{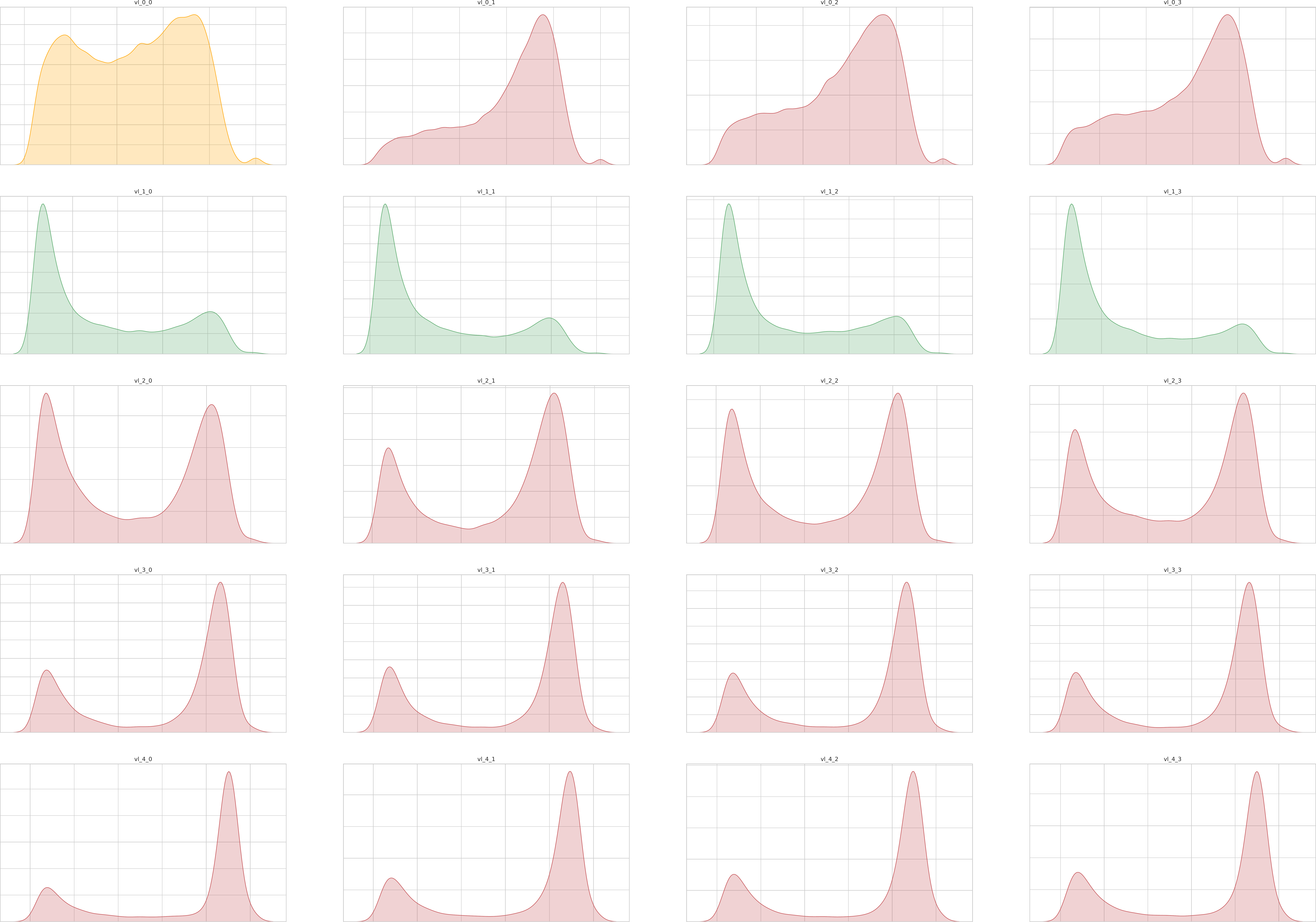}
    \end{minipage}
    &
    \begin{minipage}{3.7cm}
        \includegraphics[width=\linewidth]{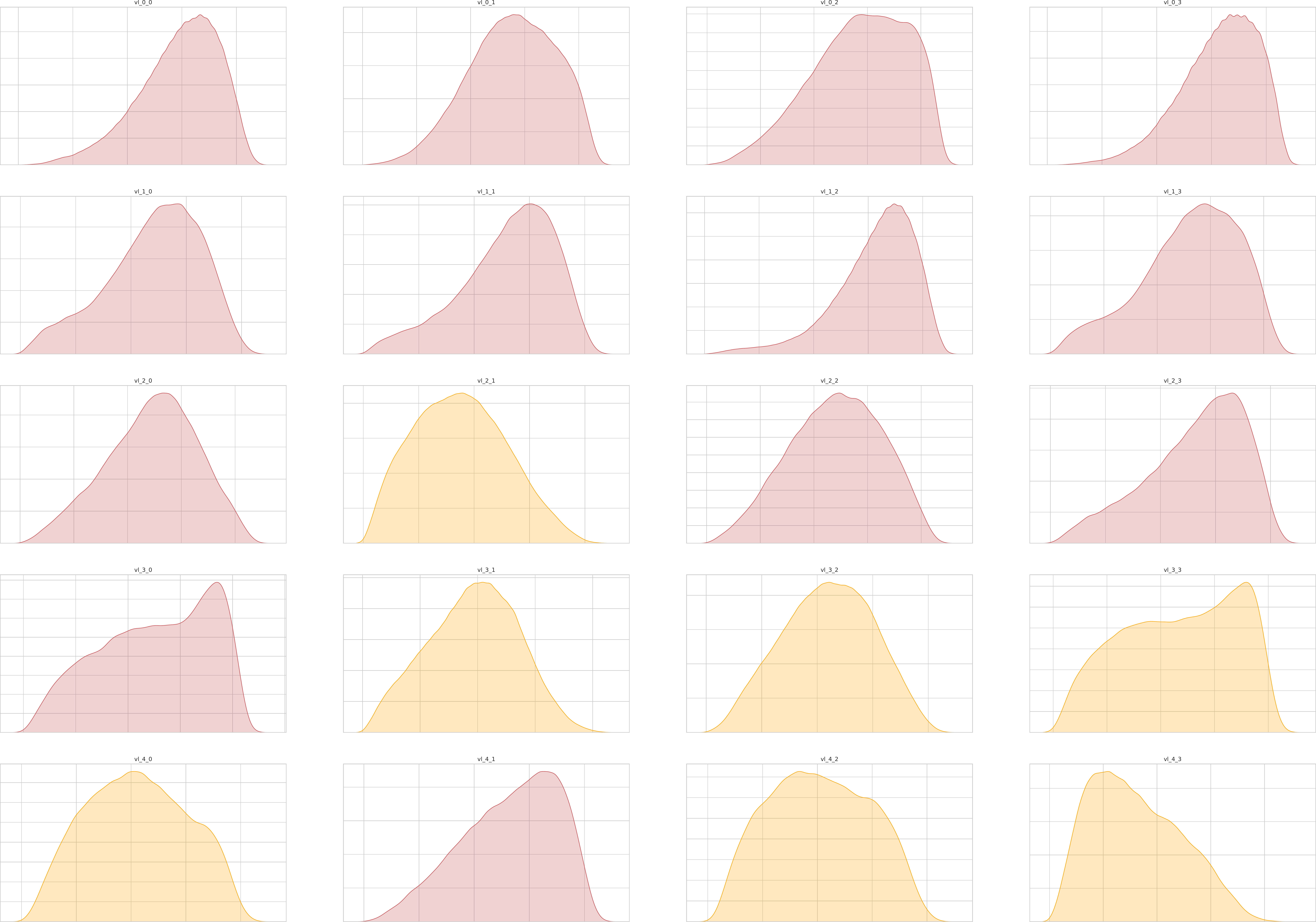}
    \end{minipage}
    \\
    (a) & (b)
    \\
\end{tabular}
}
\caption{\label{fig:vl_k_head}Comparison of $k$-distribution of $t_{\times}^{L\leftarrow V}$ attention heads for two different models: (a) oracle; (b) noisy visual input. Rows indicates different $T_{\times}^{L\leftarrow V}$ layers. Heads are colored according to the median of the $k$-number.}
\end{figure}

In contrast to the attention modes reported in~\cite{ramsauer2020hopfield}, we also observed bi-modal
$k$-number distributions, shown in Fig.~\ref{fig:attention_modes}-a, which are a combination of a dirac ( Fig.~\ref{fig:attention_modes}-b) and uniform (cf Fig.~\ref{fig:attention_modes}-c) attention modes.
We call these modes \qemph{bimorph} attention, since they reveal the existence of two different shapes of attention distribution: for some samples, a dirac activation is generated, while other samples lead to uniform attention (averaging over tokens){\color{red}\footnotemark[\value{footnote}]}.
Besides, in Fig.~\ref{fig:vl_k_head}, we compare attention mode diversity between the noisy visual model and the oracle $t_{\times}^{L\leftarrow V}$ heads, where we observe higher diversity for the oracle. In particular, \qemph{bimorph} attention are mostly performed by the oracle. 

\footnotetext{We remind that these plots are distributions of indicators of distributions: uniform behavior does not show up as a flat plot, but as plot with a peak on the right side --- it may in these plots look like a Dirac.}

\subsection{Attention modes and task functions}

\noindent
In this experiment, we study the relationships between attention modes and question types, which correspond to different functions of reasoning required to solve the problem instance. In other words, we explore to what extent the neural model adapts its attention distribution to the question at hand. We group the set of questions according to functions using the GQA~\cite{hudson2019gqa} annotation, using $54$ different functions such as \eg{} `filter color', `verify size', 
\etc{}.\footnote{There is limited overlap between functions, \eg{} `filter' contains, among others, the `filter color' and `filter size'.}.

\begin{figure}[t] 
\centering
\includegraphics[width=\linewidth]{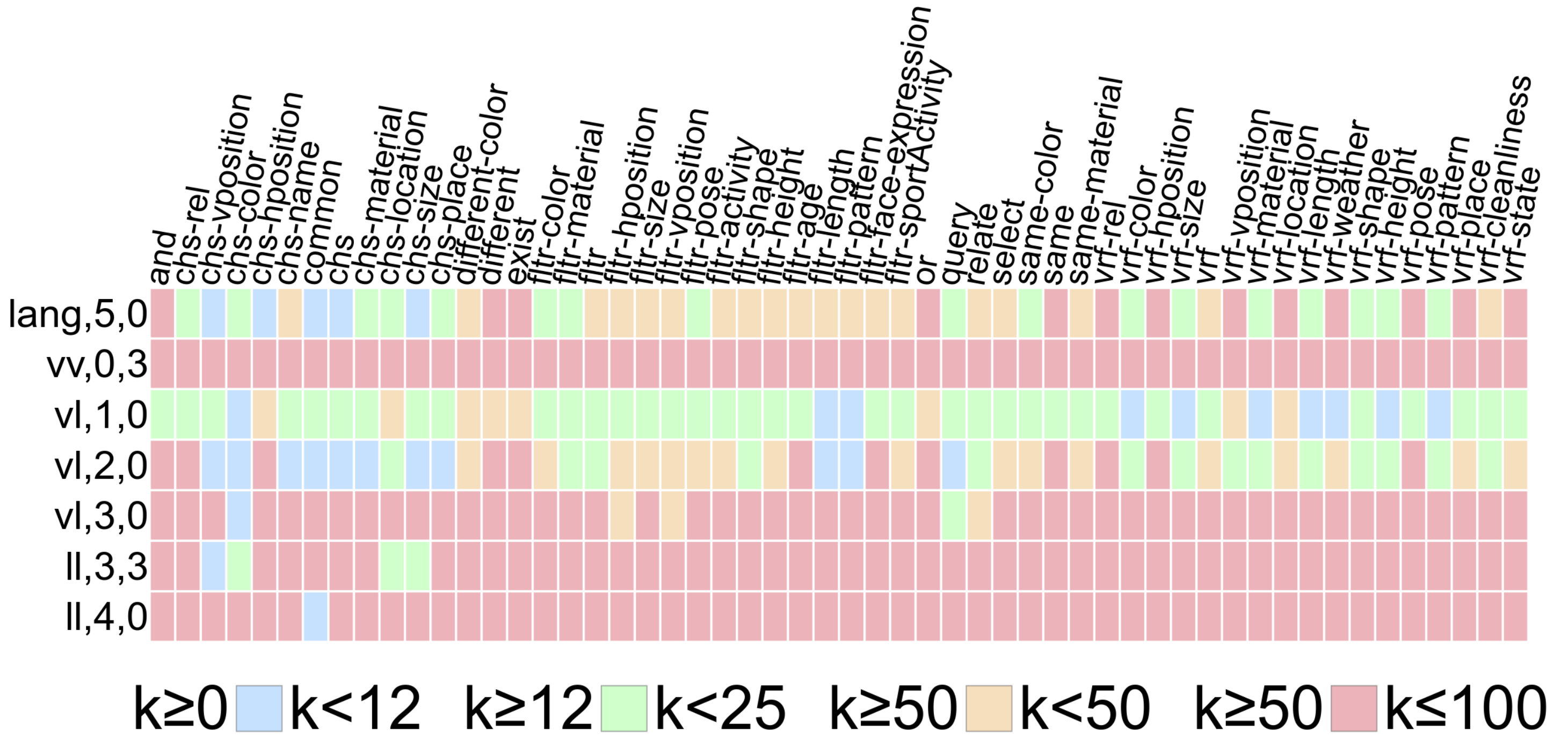}
\caption{
\label{fig:ex_heads}
Attention modes for selected attention heads (rows) related to functions required to be solved to answer a question (columns). 
The head's notation $x,i,j$ refers to the head $j$ of the $i$-th Transformer layer of type $x$: `lang'/`ll'=$t_{-}^{L}(\cdot)$, `vis'/`vv'=$t_{-}^{V}(\cdot)$, `vl'=$t_{\times}^{L\leftarrow V}(\cdot)$, `lv'=$t_{\times}^{V\leftarrow L}(\cdot)$. The VL-Transformer's architecture is presented in supp. mat..
The color encodes the attention mode, \emph{i.e} median of the $k$-number~\cite{ramsauer2020hopfield}. We observe (1) attention heads behave differently depending on the function; (2) a given function causes different attention modes for different heads.
}
\end{figure}

\begin{figure}[t] \centering
{\footnotesize
\begin{tabular}{cc}
    (a) & \multirow{2}{*}{
    \begin{minipage}{6.8cm}
        \includegraphics[width=\linewidth]{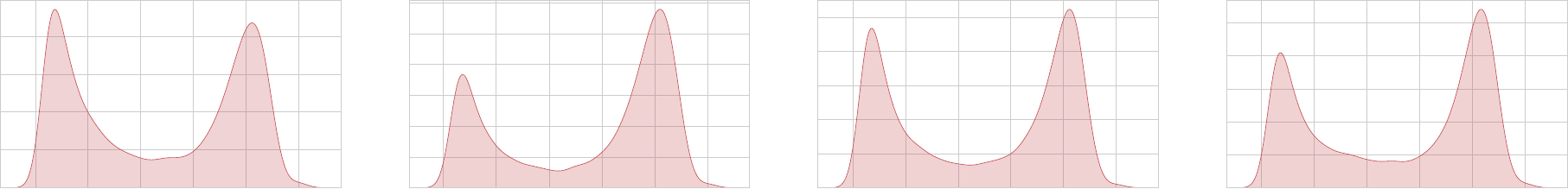}
    \end{minipage}}\\
    \small{overall} & \\
    &\\
    (b) &
    \multirow{3}{*}{\begin{minipage}{6.8cm}
        \includegraphics[width=\linewidth]{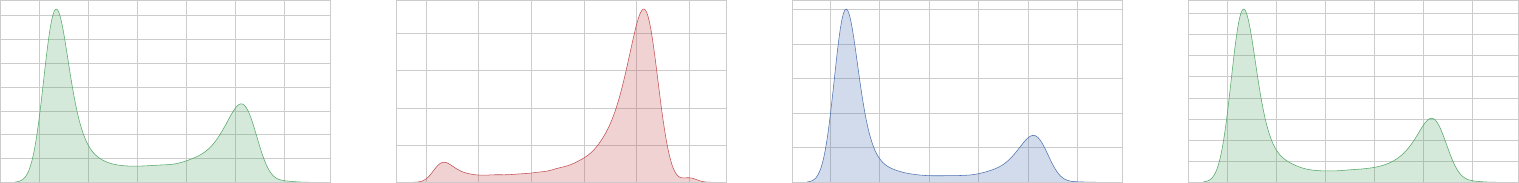}
    \end{minipage}}\\
    \small{choose} & \\
    \small{color} & \\
\end{tabular}
}
\caption{\label{fig:choose_color}Influence of the question on oracle's \qemph{bimorph} attention heads. We compare attention modes of the third layer of $T_{\times}^{L\leftarrow V}$ heads as a distribution of the $k$-numbers~\cite{ramsauer2020hopfield} over  (a) samples of all functions, and (b) samples with questions involving the `choose color' function, and observe a clear difference.
The function `choose color' seems to cause the activation (\emph{i.e} emergence of a small meta-stable state) of the $1^{st}$,  $2^{nd}$ and $4^{th}$ head, and the desactivation of the $3^{rd}$ one, further indicating  task dependence of attention head behavior.} 
\end{figure}

We link functions to the attention modes introduced in Section~\ref{sec:attention_modes}.
In Fig.~\ref{fig:ex_heads} we show functions in columns and a selection of attention heads in rows, while the color encodes the median $k$-number for the oracle model. We observe a strong dependency. 
Certain functions, \eg{} the majority of the `choose X' functions, tend to cause the emergence of small meta-stable states.
In these modes, the attention mechanism is fundamental as it allows the model to attend to specific token combinations by detecting specific patterns.
On the other hand, some functions requiring to attend to very general image properties, such as `choose location' or `verify weather', seem to be connected to very large meta-stable states. We conjecture, that to find general scene properties, a large context is needed. In this modes, the attention mechanism is less important, and replacing it with a simple averaging operation is likely to keep performance --- an experiment we explore in Section~\ref{sec:pruning}.
Similarly, when focusing on heads instead of functions, we observe that
a majority of heads typed as $t_{\times}^{V\leftarrow L}(\cdot)$ or $t_{-}^{V}(\cdot)$ tends to behave independently of the question functions and they generally show close-to-uniform attention. On the other hand, the $t_{-}^{L}(\cdot)$ and $t_{\times}^{L\leftarrow V}(\cdot)$ heads are highly dependant on the question functions. As shown in Fig.s~\ref{fig:ex_heads} and \ref{fig:choose_color}, these heads does not behave in the same way and are not `activated' (\emph{i.e} have a smaller metastable-state) for the same combination of functions. This provides some evidence for modularity of the oracle VL-Transformer, each attention head learning to specialize to one or more functions.
In addition, in Fig.~\ref{fig:choose_color}, we visualize the difference in oracle attention modes between two different function configurations: Fig.~\ref{fig:choose_color}-a is the distribution of median $k$-numbers over \emph{all} samples, \emph{i.e} involving all functions, whereas Fig.~\ref{fig:choose_color}-b shows the distribution over samples involving the `choose color' function. We show the 3rd $T_{\times}^{V\leftarrow L}$ Transformer layer heads. Over all functions, these heads show \qemph{bimorph} behavior, whereas on questions requiring to choose a color, these same heads show either dirac or uniform behavior.

\begin{figure}[t] \centering
{\footnotesize
\begin{tabular}{cc}
    \multicolumn{2}{c}{
    \begin{minipage}{7.8cm}
        \includegraphics[width=\linewidth]{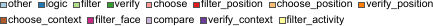}
    \end{minipage}}
    \\
    &
    \\
    \begin{minipage}{3.90cm}
        \includegraphics[width=\linewidth]{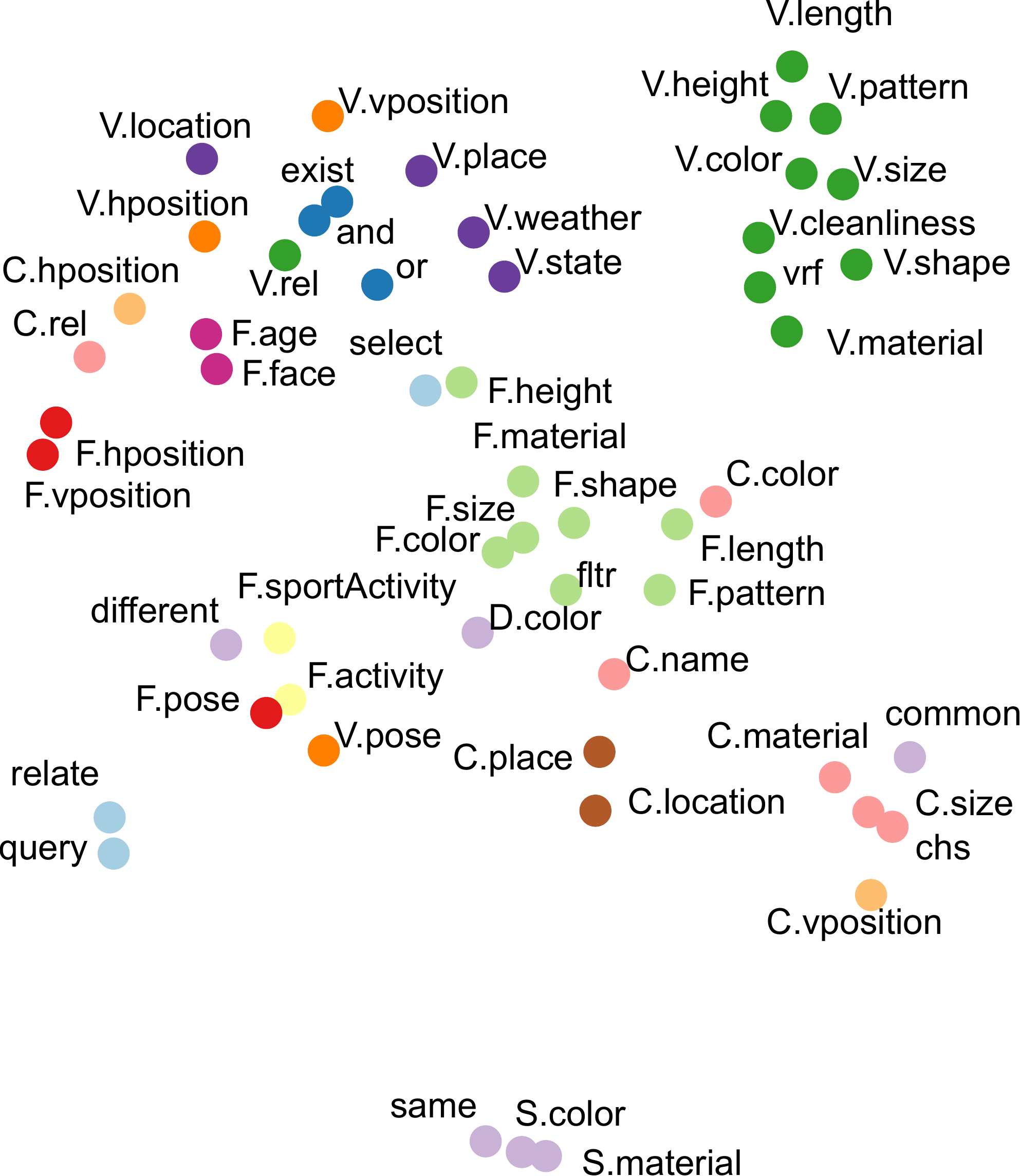}
    \end{minipage}
    &
    \begin{minipage}{3.9cm}
        \includegraphics[width=\linewidth]{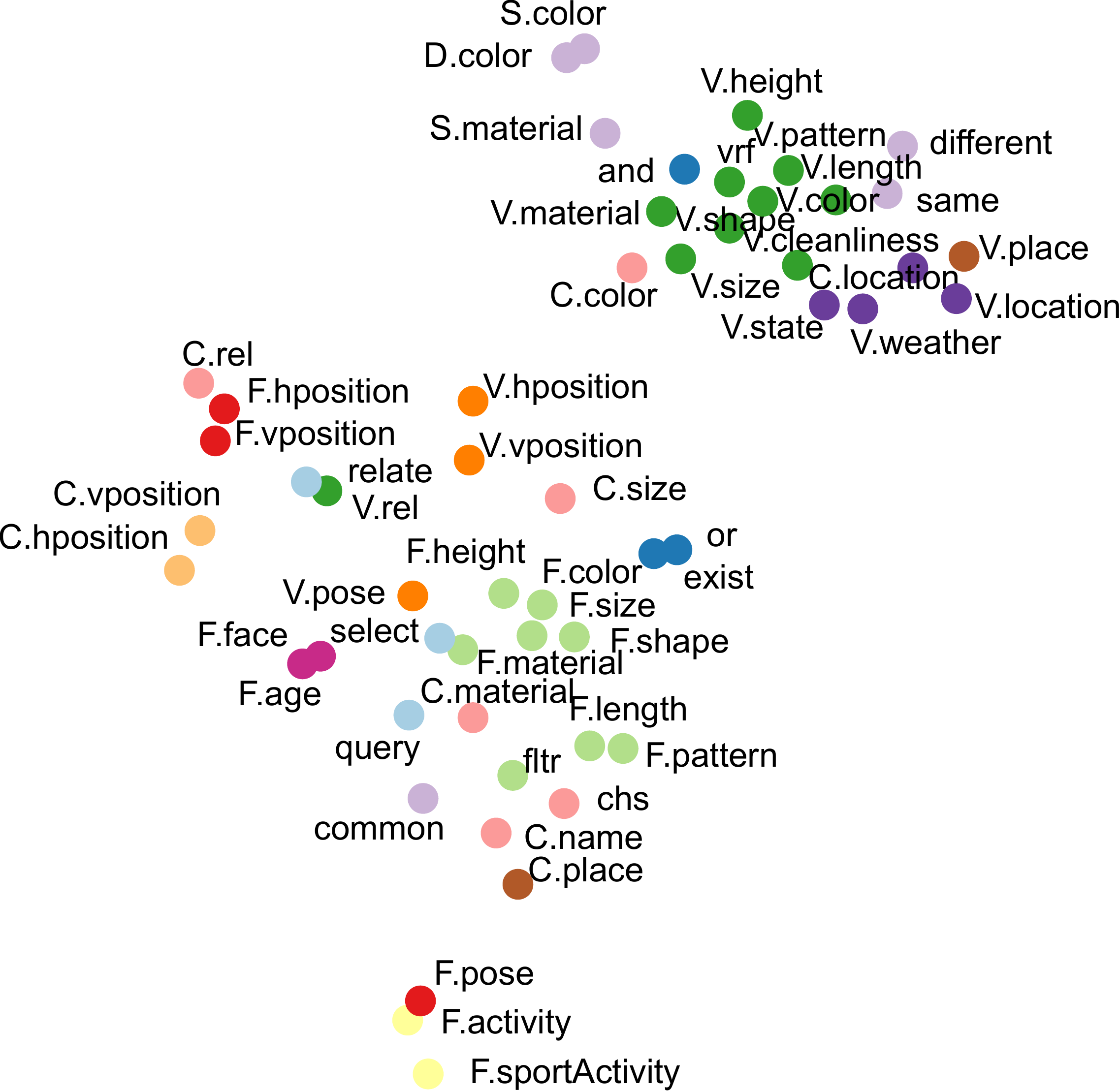}
    \end{minipage}
    \\
    (a) Oracle & (b) Noisy visual model
    \\
\end{tabular}
}
\caption{\label{fig:function_tsne}
t-SNE projection of the attention mode space, \emph{i.e} the $80$-dim representation median $k$-numbers, one per head of the model. Colors are functions, also provided as overlaid text. We compare projections of (a) the oracle, and (b) the noisy visual model, and observe a clustering of functions in the attention mode space for the oracle, but significantly less for the noisy input model.} 
\end{figure}

\myparagraph{Oracle vs. Noisy Input}
In the next experiment, we explore the difference in behavior between the perfect-sighted oracle and the classical model taking noisy visual input. For each input sample, we create a $80$-dimensional representation describing the attention behavior of the model by collecting the $k$-numbers of the $80$ cross-attention heads into a flat vector, taking the median over the tokens for a given head.
Fig.~\ref{fig:function_tsne} shows two different t-SNE projections of these attention behavior space, one for the oracle model and one for the noisy model. 
While the former produces clusters regrouping functions according to their general type, the function representation of the noisy model is significantly more entangled.
We conjecture, that the attention-function relationship provides insights into the reasoning strategies of the model. VQA requires to handle a large variety of reasoning skills and different operations on the input objects and words. Question-specific manipulation of words and objects is essential for correct reasoning. In contrast to the oracle one, the t-SNE plot for the noisy visual model paints a muddier picture, and does not show clear relationships between attention modes and  functions.

\myparagraph{Caveat} visualizing attention modes does not provide any indication of the attention operation itself, only about the shape of the operation. In particular, an attention head might result in the same low $k$-number for two different input samples, showing Dirac attention, but could attend do quite different objects or words in both cases.

\subsection{Attention pruning}
\label{sec:pruning}

\noindent
We further analyze the role of attention heads by evaluating the effect of pruning heads on model performance.
As reported by~\cite{voita2019analyzing, ramsauer2020hopfield}, specific attention heads may be useful during training, but less useful after training. In the same lines, for specific heads we replace the query-key attention map by a uniform one, \qemph{pruned} heads will therefore simply contextualize each token by an averaged representation of all other tokens, as a head with large meta-stable state would have done. 
In Table~\ref{tab:prune} we report the effect of pruning on GQA validation accuracy according to different attention categories and observe that the oracle model is resilient to pruning of the $t_{-}^{V}(\cdot)$ and $t_{\times}^{V\leftarrow L}(\cdot)$ heads, but that pruning of 
$t_{-}^{L}(\cdot)$ and $t_{\times}^{L\leftarrow V}(\cdot)$ heads results in sharp drops in performance. This indicates that the bulk of reasoning occurs over the language tokens and embeddings, which are contextualized from the visual information through  $t_{\times}^{L\leftarrow V}(\cdot)$ cross-attention. We can only conjecture why this solution emerges after training --- we think that among reasons are the deep structure of language and the fact that in current models the answer is predicted from the CLS language token.

\begin{table}[] \centering
{\small
    \begin{tabular}{lccccc}
    \toprule
        Pruned attentions & n/a & L & V & L$\leftarrow$V & V$\leftarrow$L \\
    \midrule
        Accuracy & 91.5 & 37.9 & 91.4 & 52.8 & 68.1 \\
    \bottomrule
    \end{tabular}
    }
    \caption{\label{tab:prune}Impact of pruning different types of attention heads of the trained oracle model. We observe that `vision' and `language$\rightarrow$vision' Transformers are hardly impacted by pruning, in contrast to `language' and `vision$\rightarrow$language'.
    Accuracies (in $\%$) on the GQA validation set.}
\end{table}

\begin{figure}[t] \centering
\begin{tabular}{cc}
    \multicolumn{2}{c}{
    \begin{minipage}{7cm}
        \includegraphics[width=\linewidth]{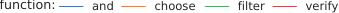}
    \end{minipage}}
    \\
    \begin{minipage}{3.5cm}
        \includegraphics[width=\linewidth]{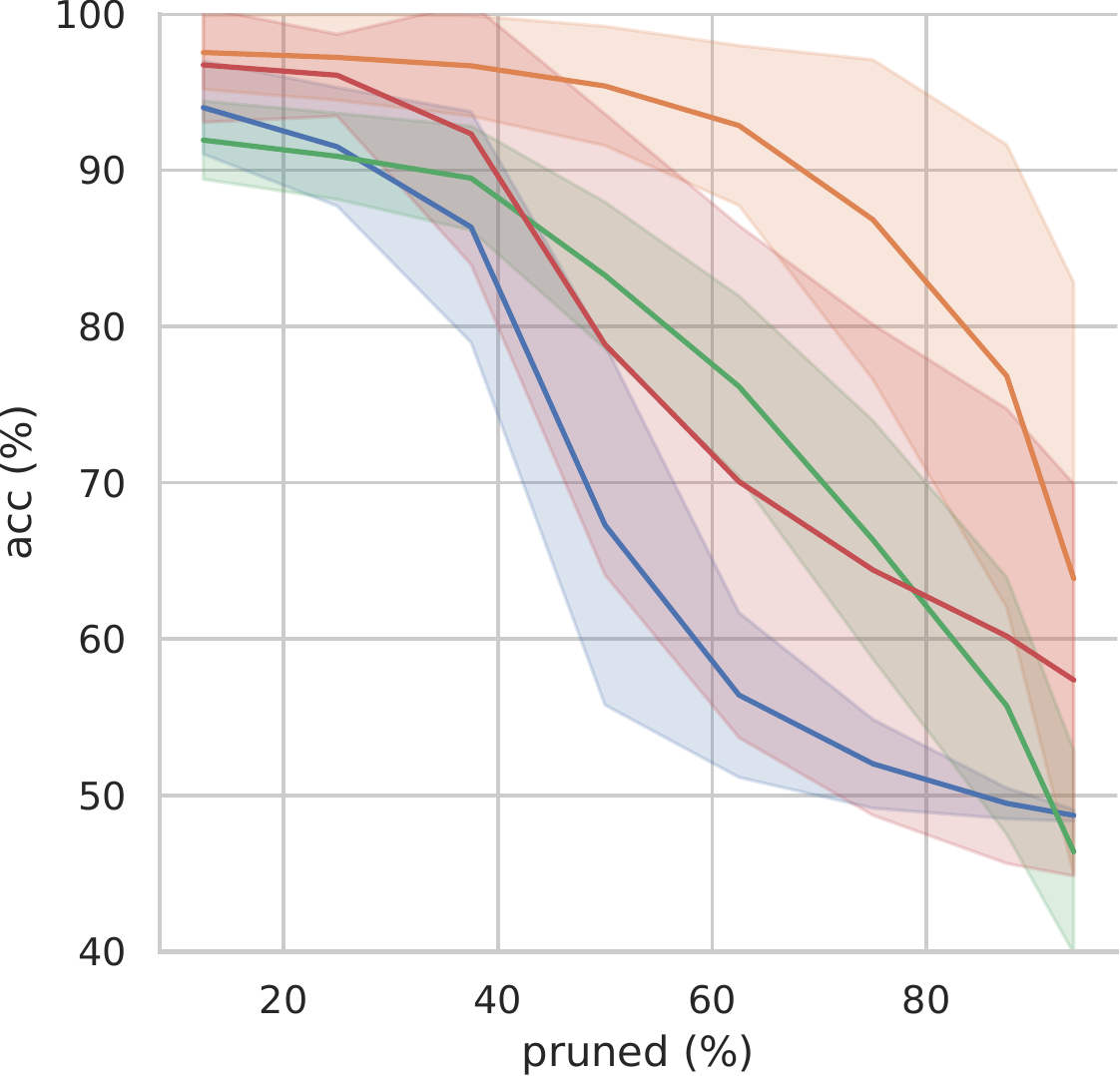}
    \end{minipage}
    &
    \begin{minipage}{3.5cm}
        \includegraphics[width=\linewidth]{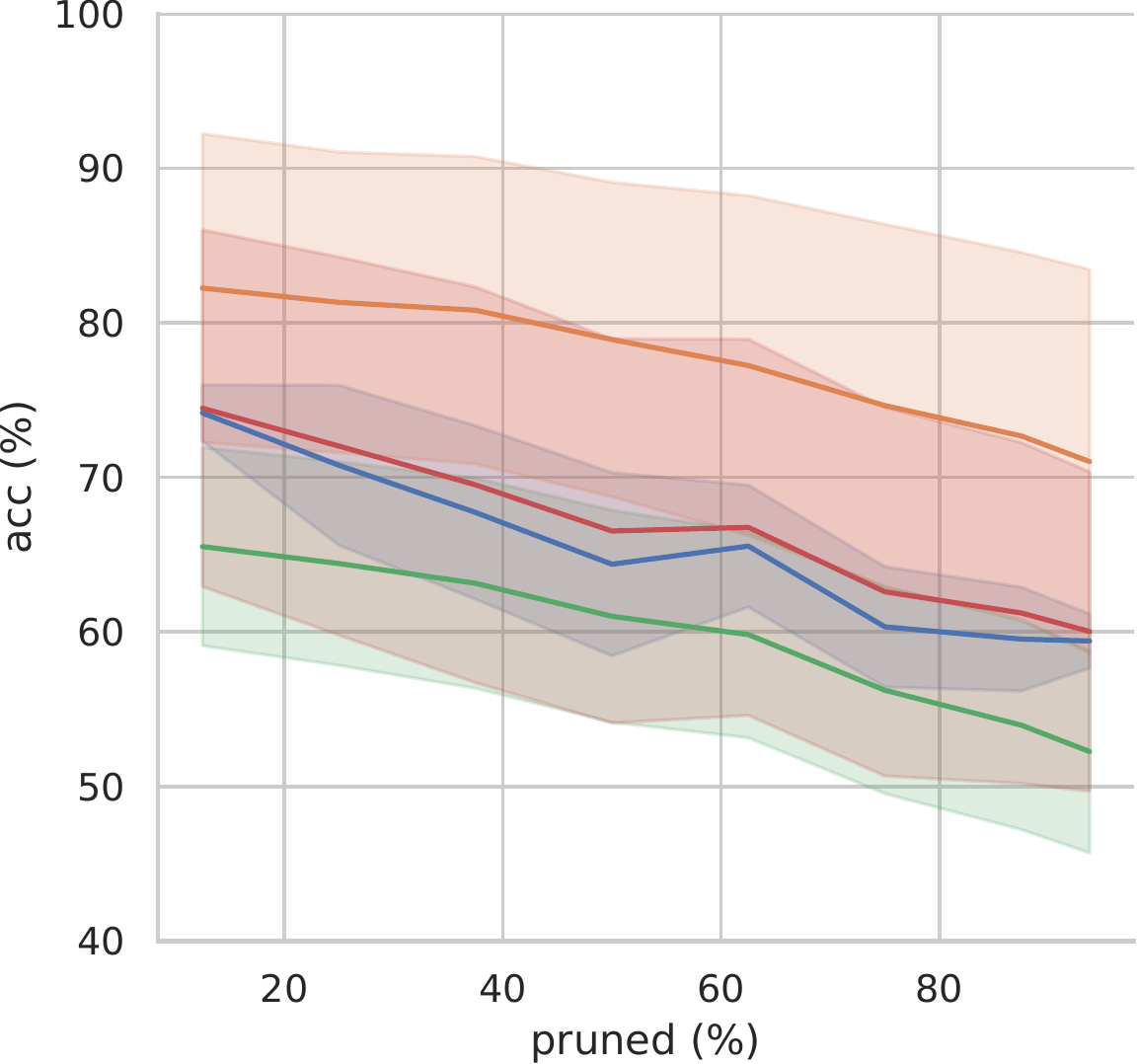}
    \end{minipage}
    \\
    (a) Oracle & (b) Noisy visual 
    \\
\end{tabular}
\caption{\label{fig:pruning}Impact of random pruning of varying numbers of attention heads in cross-modal layers on GQA-validation accuracy. (a) For the oracle, the impact is related to the nature of the function, highlighting its modular property. 
(b) For the noisy-vision-based model, pruning seems to be unrelated to function types.}
\end{figure}

\myparagraph{Impact on functions}
We study the impact of pruning on the different task functions by randomly pruning $n$ cross-attention heads and measuring accuracy for different function groups, n being varied between $0\%$ (no pruning) to $100\%$ (all heads are pruned), as shown in Fig.~\ref{fig:pruning} for the oracle and noisy vision-based models. For the sake of clarity only $4$ different function are shown,  additional results are provided in supplementary material. 
For the perfect-sighted oracle (Fig.~\ref{fig:pruning}-a), we first observe that the pruning has a different impact depending on the function. Thereby, while \emph{filter} and \emph{choose} are dominated by negative curvature where performance drops only when a large number of heads are pruned, \emph{verify} and \emph{and}, are characterized by a sharp inflection point and an early steep drop in performance. This indicates that the model has learned to handle functions specifically, resulting in various degrees of reasoning distribution over attention heads.
For the noisy vision-based model, on the other hand, the effect of head pruning seems to be unrelated to the function type (Fig.~\ref{fig:pruning}-b). 

\begin{figure*}[t] \centering
\includegraphics[width=\linewidth]{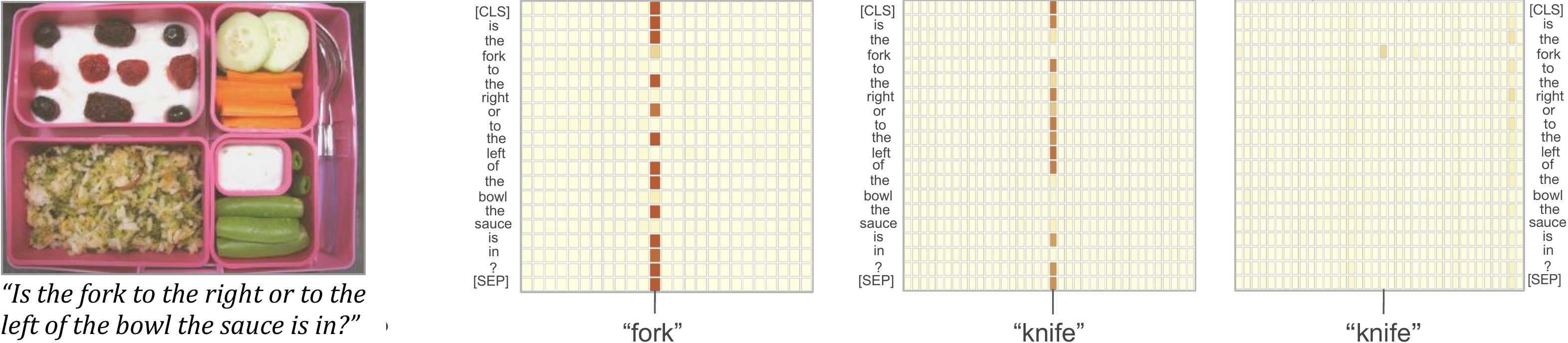} \\
\hspace{50mm} (a) Oracle \hspace{20mm} (b) Oracle transfer \hspace{10mm} (c) Baseline 
\caption{\label{fig:theotool}
Example for the difference in attention in the second $T_{\times}^{L\leftarrow V}$ layer. The oracle drives attention  towards a specific object, ``fork'', also  seen after transfer but not in the baseline (we checked for permutations). The transferred model overcame a miss-labelling of the fork as a knife. This analysis was performed with our interactive visualization tool, which also allows to visualize attention models, not shown here ({\bf \footnotesize{\toolUrl}}, online experience + source code; video provided in the supp. mat.).
}
\end{figure*}

\subsection{Interactive visualization}

\noindent
The analysis described above was based on integrating information of various kinds over a full dataset, GQA-validation. Additional insights can be gained by exploring the behavior of individual problem instances and relating them to statistics extracted from the population, in particular attention modes and groups of functions. We have performed this analysis on a large number of samples and we provide a tool, which allows the reader to perform similar experiments online, making it possible to load different oracle or noisy vision based models. This tool is available at {\color{red}\footnotemark[1]} (online experience + source code). Fig.~\ref{fig:theotool} gives a simple visualization (see Section~\ref{sec:transferringpatterns} for a discussion), a video of its usage is provided in the supplementary material.

\myparagraph{Discussion}
The experiments of this section have shown a pronounced difference in attention modes between the perfectly-sighed oracle and a noisy vision based model. More importantly, the oracle model shows a strong relationship between attention mode and task function, which we interpret as the capability of adapting reasoning to the task at hand. The classical model significantly lacks this abilities, suggesting a strategy of transferring patterns of reasoning from an oracle model pre-trained on visual GT to a model taking noisy visual input.

% -------------------------------------------------------------------------
\section{Transferring Reasoning Patterns}
\label{sec:transferringpatterns}

\noindent
We propose \textit{Oracle Transfer}, transferring reasoning patterns from a perfectly-sighted model to a deployable model taking noisy visual inputs. We argue, that the first optimization steps are crucial for the emergence of specific attention modes. Training proceeds as follows (see Fig. \ref{fig:teaser}):
\begin{enumerate}
    \item Training of a perfectly-sighted oracle model on GT visual inputs from the GQA~\cite{hudson2019gqa} annotations, in particular a \emph{symbolic} representation concatenating the 1-in-K encoded object class and attributes of each object.
    \item {Initialize a new model \emph{with the oracle parameters}}. This new model is taking noisy visual input in a form of the \emph{dense} representation (2048-dim feature vector extracted by Faster-RCNN~\cite{ren2015faster} fused with bounding-boxes). The first visual layers ($T_{-}^{V}$) are initialized randomly due to the difference in nature between dense and symbolic representations.
    \item Optionally and complementary, \emph{continue training} with large-scale self-supervised objectives (LXMERT~\cite{tan2019lxmert}/BERT-like) on combined data from Visual Genome~\cite{krishna2017visual}, MS COCO~\cite{lin2014microsoft}, VQAv2~\cite{goyal2017making}.
    \item {\emph{Fine-tune}} with the standard VQA classification objective on the target dataset (GQA~\cite{hudson2019gqa} or VQAv2~\cite{goyal2017making}).
\end{enumerate}

\myparagraph{Experimental setup}
We use the same VL-Transformer architecture defined in Section~\ref{sec:vis_attention} (more details in supp. mat.), with $d{=}128$ and $h{=}4$, which corresponds to a tiny version of LXMERT~\cite{tan2019lxmert} architecture.
Following \cite{tan2019lxmert}, we use $36$ objects per image.  
We evaluate on the GQA~\cite{hudson2019gqa}, GQA-OOD~\cite{RosesAreRed} and VQAv2~\cite{goyal2017making} datasets. 
\begin{figure}[t] \centering
{\footnotesize
\begin{tabular}{cc}
    (a) & \multirow{2}{*}{
    \begin{minipage}{6.2cm}
        \includegraphics[width=\linewidth]{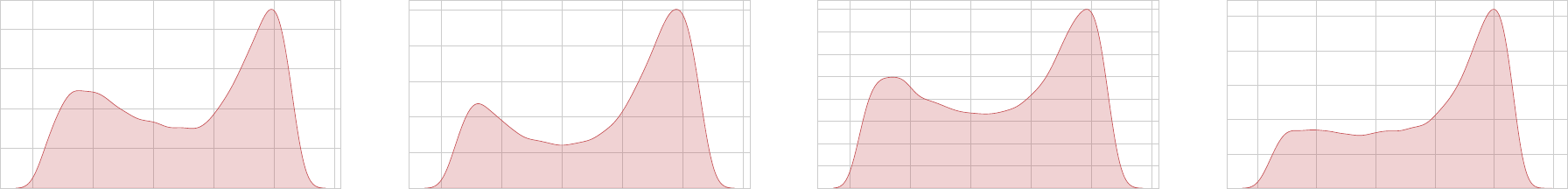}
    \end{minipage}}\\
    \small{overall} & \\
    &\\
    (b) &
    \multirow{3}{*}{\begin{minipage}{6.2cm}
        \includegraphics[width=\linewidth]{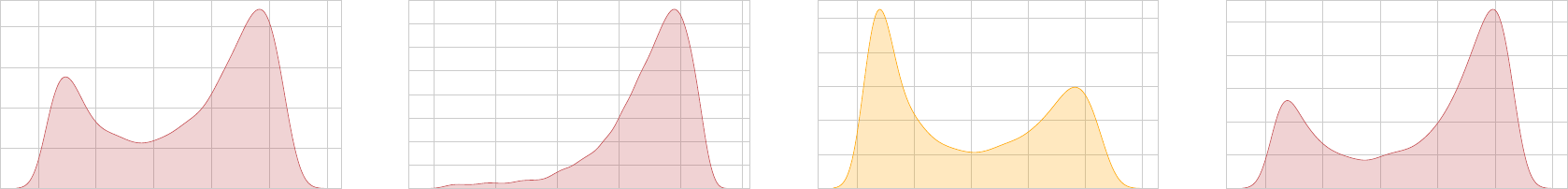}
    \end{minipage}}\\
    \small{choose} & \\
    \small{color} & \\
\end{tabular}
}
\caption{\label{fig:transfer_choose_color} We reproduce Fig.~\ref{fig:choose_color} with our VL-Transformer + dense \emph{Oracle Transfer} (same heads/layers). As we can see in (a), the attention heads have retained their \qemph{bimorph} property, although their shape is distorted by the noisy visual training. In addition, when we measure the attention mode on questions involving the choose color function, in (b), we observe that the attention heads are still function-dependant, although in a lesser extent.} 
\end{figure}

\begin{table*}[] \centering
{\small
    \begin{tabular}{l|cc|cc|c|c}
        \toprule
        \multirow{2}{*}{Model} & \multicolumn{2}{c|}{Pretraining} &  \multicolumn{2}{c|}{GQA-OOD}~\cite{RosesAreRed} & GQA~\cite{hudson2019gqa} & VQAv2~\cite{goyal2017making}\\
        & Oracle & LXMERT/BERT & acc-tail & acc-head & overall & overall\\
        \midrule
        (a) Baseline & & &  42.9 & 49.5 & 52.4 & - \\
        \textbf{(b)} Ours & \checkmark & &   \textbf{48.5} & \textbf{55.5} & \textbf{56.8} & - \\
        \midrule
        (c) Baseline (+LXMERT/BERT) & & \checkmark &  47.5 & 54.7 & 56.8 & 69.7\\
        \textbf{(d)} Ours (+LXMERT/BERT) & \checkmark &  \checkmark & \textbf{48.3} & \textbf{55.2} & \textbf{57.8} & \textbf{70.2}\\
        \bottomrule
    \end{tabular}
}
\vspace*{1mm}
\caption{\label{tab:oracle_transfert_tiny} Quantitative evaluation of the proposed knowledge transfer from oracle models. All listed models are deployable, no GT input is used for testing. Models: (c)+(d) are pre-trained with LXMERT~\cite{tan2019lxmert}/BERT-like objectives after \emph{Oracle Transfer}. 
All scores GQA-OOD-testdev~\cite{RosesAreRed}; GQA~\cite{hudson2019gqa}-testdev; VQAv2-test-std~\cite{goyal2017making}. Training hyperparameters selected on respective validation sets.}
\end{table*}

\myparagraph{Evaluating transfer}
We evaluate the impact of \emph{Oracle Transfer} on three different benchmarks in Table~\ref{tab:oracle_transfert_tiny}, observing that transferring knowledge from the oracle significantly boosts accuracy. We also evaluate the effect of \emph{Oracle Transfer} on bias reduction and benchmark on GQA-OOD~\cite{RosesAreRed}, reporting gains in Out-Of-Distribution settings --- rare samples, ``acc-tail'' --- by a large margin, which suggests improved generalization ability.
Our experiments show that \emph{Oracle Transfer} is complementary to large-scale vision-language self-supervised objectives of type LXMERT/BERT-like pretraining as introduced in~\cite{tan2019lxmert}. An overall gain of about $+1$ accuracy points is observed from models (c) to (d) in Table~\ref{tab:oracle_transfert_tiny}, attributed to \emph{Oracle Transfer}. As a comparison, LXMERT/BERT pretraining alone does not improve ``acc-tail'' on GQA-OOD.

\myparagraph{Cross-dataset training}
We explore whether the effects of oracle knowledge generalize beyond the GQA dataset, and evaluate training the oracle on GQA GT annotations, performing LXMERT/BERT pretraining, and transferring to a model trained on VQAv2 dataset~\cite{goyal2017making}. We improve VQAv2 accuracy by a significant margin, suggesting positive transfer beyond GQA (Table~\ref{tab:oracle_transfert_tiny}).

\begin{table}[] \centering
{\small
    \begin{tabular}{lccc}
        \toprule
        Method & Input train & Input test &  Acc.\\
        \midrule
        (a) Baseline & Dense & Dense & 61.7 \\
        (b) Transf. w/o retrain & 1-in-K GT & 1-in-K pred. & 58.8 \\
        (c) Transf. w/ $T_{-}^{V}$ retrain & 1-in-K GT & Dense & 61.7 \\
        (d) Transf. w/ retrain &1-in-K GT & Dense & 66.3 \\
        \bottomrule
    \end{tabular}
}
\caption{\label{tab:ablation_transfert} Impact of different types of transfer, GQA~\cite{hudson2019gqa} val. accuracy. All models are deployable (no GT used for testing).}
\end{table}

\myparagraph{Transfer ablation studies}
We evaluate different variants of knowledge transfer, shown in Table~\ref{tab:ablation_transfert}, on the GQA validation set only. We explore a direct transfer from the oracle to a deployable model without retraining, by making visual input representations comparable. To this end, the deployable model receives 1-in-K encoded class information, albeit not from GT classes but taking classes from the Faster R-CNN detector (Table~\ref{tab:ablation_transfert}-b). While inferior to the baseline, its performance is surprisingly high, suggesting that the oracle learns knowledge which is applicable in real/noisy settings. Performance gains are, however, only obtained by finetuning the model to the uncertainties in dense visual embeddings. Retraining only the visual block (Table~\ref{tab:ablation_transfert}-c), performances are on par with the baseline, retraining the full model (Table~\ref{tab:ablation_transfert}-d) gains $+4.6p$.

\myparagraph{Comparison with SOTA}
\emph{Oracle Transfer} allows to improve performance of the tiny-LXMERT model both in and out of distribution ~\cite{RosesAreRed} (Table~\ref{tab:sota}, bottom part).
Transfer is parameter efficient and achieves on-par overall accuracy with MCAN-6~\cite{kim2018bilinear} while halving capacity. 

\myparagraph{Qualitative analysis \& interpretability}
Finally, we qualitatively study the effects of \emph{Oracle Transfer} and interpretability of attention heads. As shown in Fig.~\ref{fig:transfer_choose_color}, after transfer, the VL-Transformer preserves the \qemph{bimorph} property of its attention heads, which was present in the original oracle model (Fig.~\ref{fig:vl_k_head}-a), but  absent in the baseline (Fig.~\ref{fig:vl_k_head}-b).
In addition, Fig.~\ref{fig:theotool} shows the attention maps of the $T_{\times}^{L\leftarrow V}$ heads in the second cross-modal layer for an instance. This head, referenced as $VL,1,0$ in Fig.~\ref{fig:ex_heads}, is observed to be triggered to questions such as ``verify attr'' and ``verify color'' provided as example. We observe that the oracle model draws attention towards the object ``fork'' in the image, and also, to a lesser extend, in the transferred model, but not in the baseline model. Similar attention patterns were observed on multiple heads in the corresponding cross-modal layer --- this analysis took into account possible permutations of heads between models. Interestingly the miss-classification as a ``knife'' prevents the baseline from drawing attention to it, but not the transferred model.

\begin{table}[t] \centering
{\small
    \begin{tabular}{lccccc}
    \toprule
        Method& $|\Theta|$ & O & L & OOD & GQA \\
    \midrule
        BUTD~\cite{anderson2018bottom} & 22 & & & 42.1 &  51.6 \\
        BAN-4~\cite{kim2018bilinear} & 50 & & & 47.2 & 54.7 \\
        MCAN-6~\cite{yu2019deep} & 52 & & & 46.5 & 56.3 \\
        \textbf{Ours} & 26 & \checkmark & & \textbf{48.5} & \textbf{56.8} \\
        \midrule
        LXMERT-tiny & 26 & & \checkmark & 47.5 & 56.8 \\
        LXMERT-tiny + Ours & 26 & \checkmark & \checkmark & 48.3 & 57.8 \\
        LXMERT~\cite{tan2019lxmert} & 212 & & \checkmark & 49.8 & 59.6 \\
        \midrule
        \multicolumn{6}{l}{\footnotesize{$|\Theta|$ = number of parameters (M); 
        OOD = GQA-OOD~\cite{RosesAreRed} Acc-tail.}} \\
        \multicolumn{6}{l}{\footnotesize{O = \textit{Oracle Transfer}, L = LXMERT/BERT pretraining.}} \\
        \bottomrule
    \end{tabular}
}
    \vspace*{1mm}
    \caption{\label{tab:sota}Comparison with SOTA on GQA and GQA-OOD on testdev. Hyperparameters were optimized on GQA-validation.}
\end{table}

\section{Conclusion}

\noindent
We have provided a deep analysis and visualizations of several aspects of deep VQA models linked to reasoning on the GQA dataset. We have shown, that oracle models produce significantly better results on questions with rare GT answers than models on noisy data, that their attention modes are more diverse and that they are significantly more dependent on questions. We have also performed instance level analysis and we propose a tool available online{\color{red}\footnotemark[1]}, which allows to visualize attention distributions and modes, and their links to task functions and dataset wide statistics.

Drawing conclusions from this analysis, we have shown that reasoning patterns can be partially transferred from oracle models to SOTA VQA models based on Transformers and BERT-like pre-training. The accuracy gained from the transfer is particularly high on questions with rare GT-answers, suggesting that the knowledge transferred is related to reasoning, as opposed to bias exploitation.

{\small
\textbf{Acknowledgements ---} C. Wolf acknowledges support from ANR through grant ``\emph{Remember}'' (ANR-20-CHIA-0018).
}

{\small
\bibliographystyle{ieee_fullname}
\bibliography{main}
}

\clearpage
\newpage
\appendix

\section{Interactive visualization of reasoning patterns}

\noindent
In the video added to this supplementary material, we provide a detailed analysis of the differences in attention modes for a given instance between the oracle model, the noisy baseline, and the oracle transfer model, providing indications for computer vision being the bottleneck in learning, and showing how patterns of attention are adapted by the transfer.

This video has been shot as a commented screencast of an interactive application which we made available online at \toolUrl{}. The tool has been designed to explore the behavior of individual problem instances. We hope it will help scientists to better understand attention mechanisms at work in  VL-Transformers of VQA models.

\begin{figure*}[] \centering
\includegraphics[width=\linewidth]{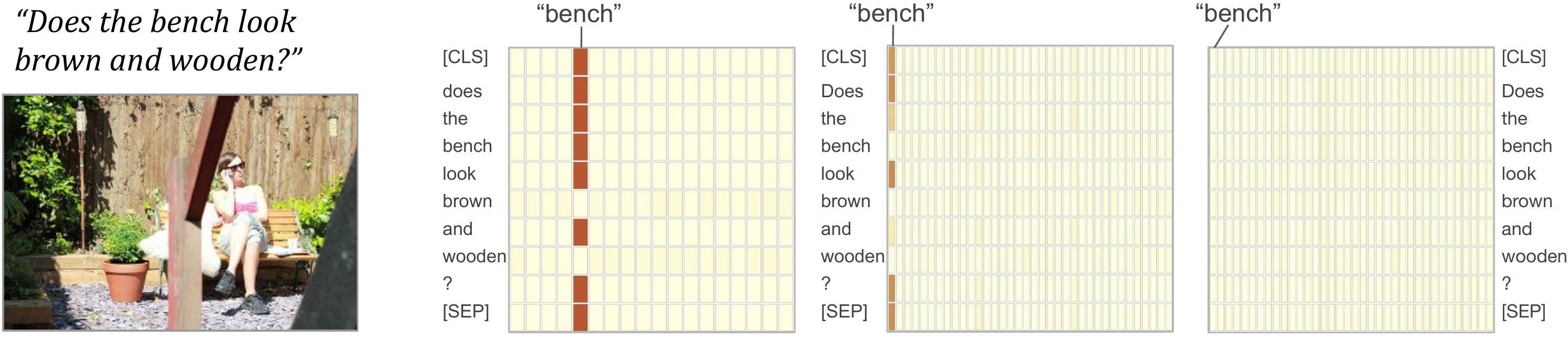} \\
\hspace{50mm} (a) Oracle \hspace{20mm} (b) Oracle transfer \hspace{10mm} (c) Baseline 
\caption{\label{fig:theotool_2}
Example for the difference in attention in the first vision to language layer. The oracle drives attention  towards a specific object, ``bench'', also  seen after transfer but not in the baseline (we checked for permutations). This analysis was performed with our interactive visualization tool, which also allows to visualize attention models, not shown here ({\bf \footnotesize{\toolUrl}})
}
\end{figure*}

In addition, we provide another example of the differences in attention in Fig~\ref{fig:theotool_2}.

\cam{
\section{Is there a confounding factor in GQA-OOD}
\noindent
Could there be a confounding factor, in which the rare answers involve objects that are simply more difficult to recognize ? Rare objects certainly had fewer visual examples for training the visual recognition models, and/or could be generally smaller or more occluded, for example.
We evaluated this by comparing the performance of the object detector for two different sets (in Table~\ref{tab:recall}):
(1) Objects required to answer questions w/ rare GT answers
(tail); and (2) objects required to answer questions w/ frequent
GT answers (head). We report similar performance, and hence no evidence supporting the hypothesis of a confounder.}

\begin{table}[] \centering
    {\footnotesize
    \begin{tabular}{cccc}
        \toprule
        GQA-OOD val. split & R@0.2 & R@0.5 & R@0.8 \\
        \midrule
        Head & 89.7\% & 77.1\% & 12.7\%\\
        Tail & 89.0\% & 75.8\% & 12.6\%\\
        \bottomrule
    \end{tabular}
    }
    \vspace*{1mm}
    \caption{\label{tab:recall} Are there confounding factors? We report R-CNN recall (R) on objects required for reasoning w/ various IoU tresholds.}
\end{table}

\begin{figure*}[th!] \centering
{
\includegraphics[width=\linewidth]{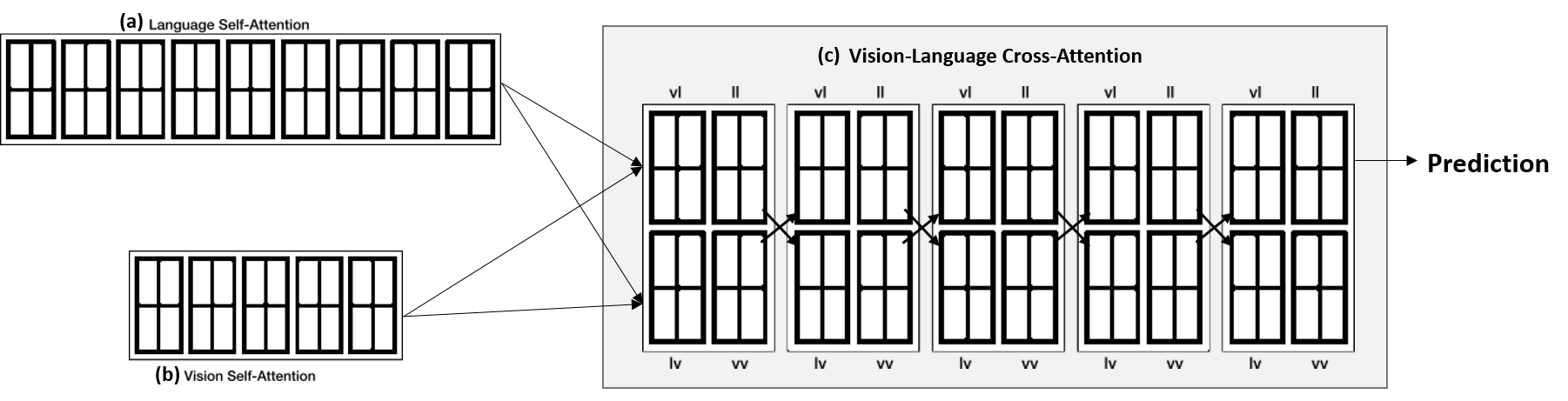}
\caption{\label{fig:vl_transformer} Schematic illustration of the VL-Transformer architecture used in the paper. It is composed of: (a) 9 $\times$ language only self-attention layers (b) 5 $\times$ vision only self-attention layers; (c) 5 $\times$ bi-directional vision-language cross-attention layers.
Crossed arrows symbolize the cross modal information flow. Small rectangles illustrate the individual attention heads. Notations and abbreviations are summarized in Table~\ref{tab:arch}.}
}
\end{figure*}

\section{Additional visualizations}

\begin{figure*}[t] \centering
\begin{tabular}{ccc}
    (a) Oracle 
    &
    \begin{minipage}{11cm}
        \includegraphics[width=\linewidth]{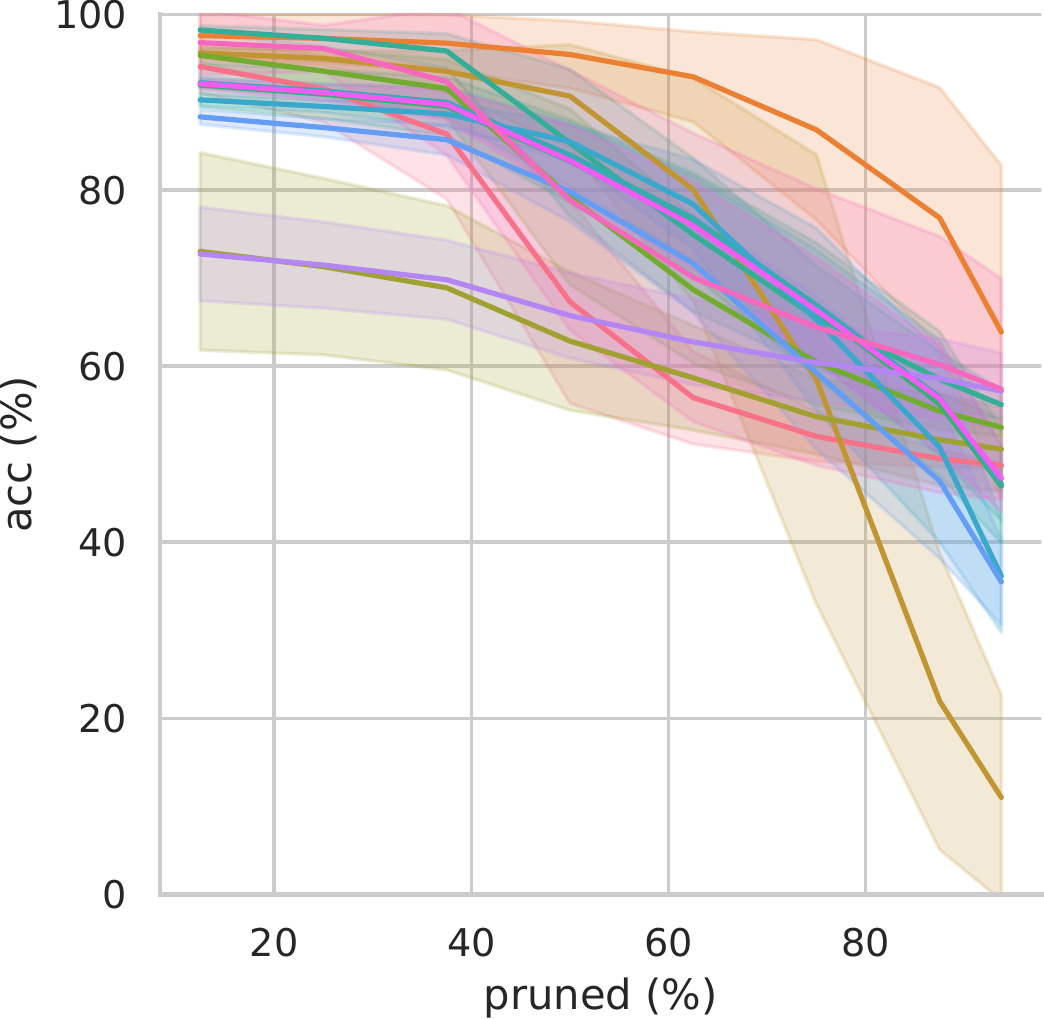}
    \end{minipage}&
    \begin{minipage}{2cm}
        \includegraphics[width=\linewidth]{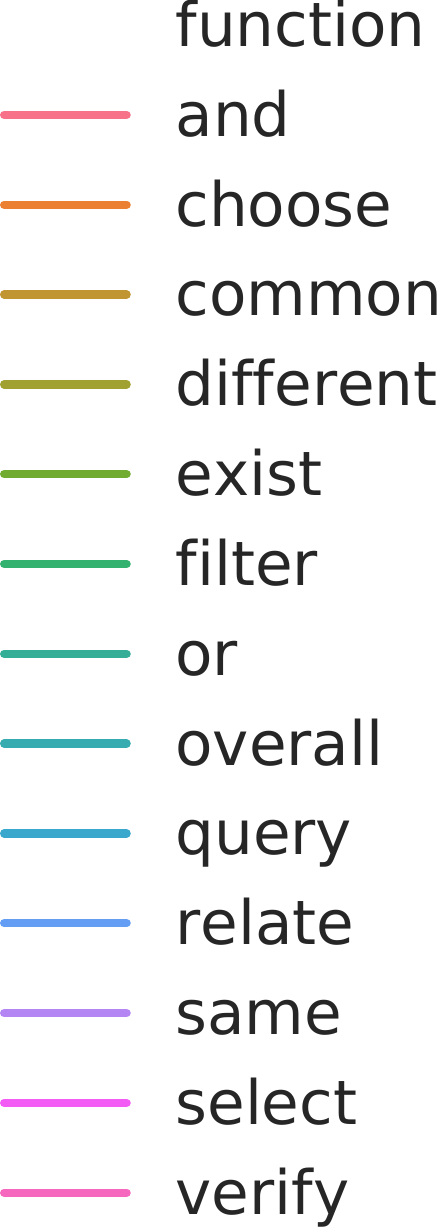}
    \end{minipage}
    \\
    (b) Noisy visual
    &
    \begin{minipage}{11cm}
        \includegraphics[width=\linewidth]{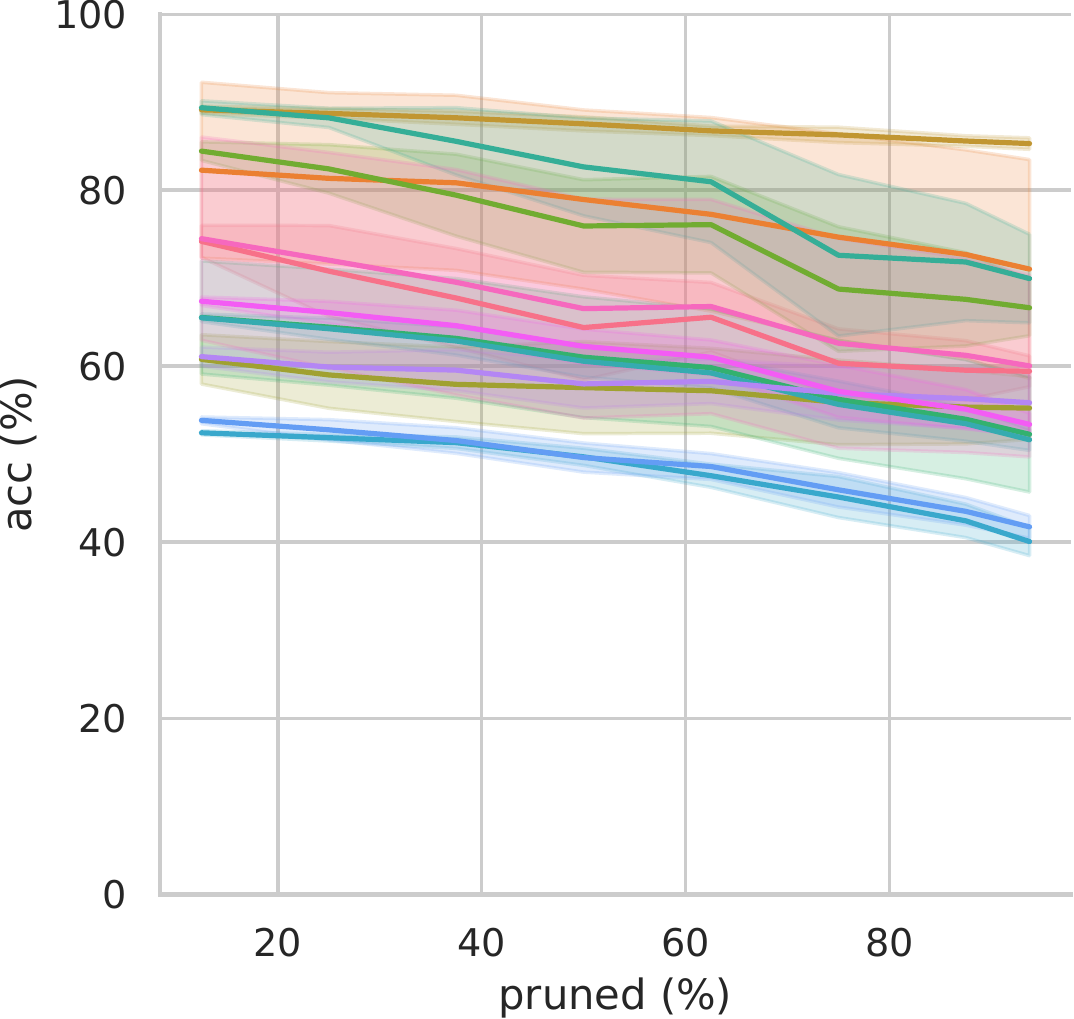}
    \end{minipage}&
    \begin{minipage}{2cm}
        \includegraphics[width=\linewidth]{images/legend_pruning_all.pdf}
    \end{minipage}
    \\
\end{tabular}
\vspace*{1mm}
\caption{\label{fig:more_pruning}Impact of random pruning of varying numbers of attention heads in cross-modal layers on GQA-validation accuracy. (a) For the oracle, the impact is related to the nature of the function, highlighting its modular property. We plot the mean and standard deviation for each function.} 
\end{figure*}

\myparagraph{Attention pruning}
We provide additional plots of the impact of attention pruning on performances, structured according to task functions, in Fig~\ref{fig:more_pruning}. Functions are gathered according to their general type, \eg{} all questions involving to ``filter something'' (\textit{filter color}, \textit{filter material}, \etc{}) are grouped as \textit{filter} function.
We still observe a significant difference between oracle and noisy visual models. In particular, for the oracle, the \textit{common} function is highly impacted by pruning. We later found that the $t_{-}^{L}(\cdot)$ heads at works in cross modal layers were essential for this function.

\begin{figure*}[t] \centering
{\footnotesize
\begin{tabular}{cc}
    (a) & \begin{minipage}{14cm}
        \includegraphics[width=\linewidth]{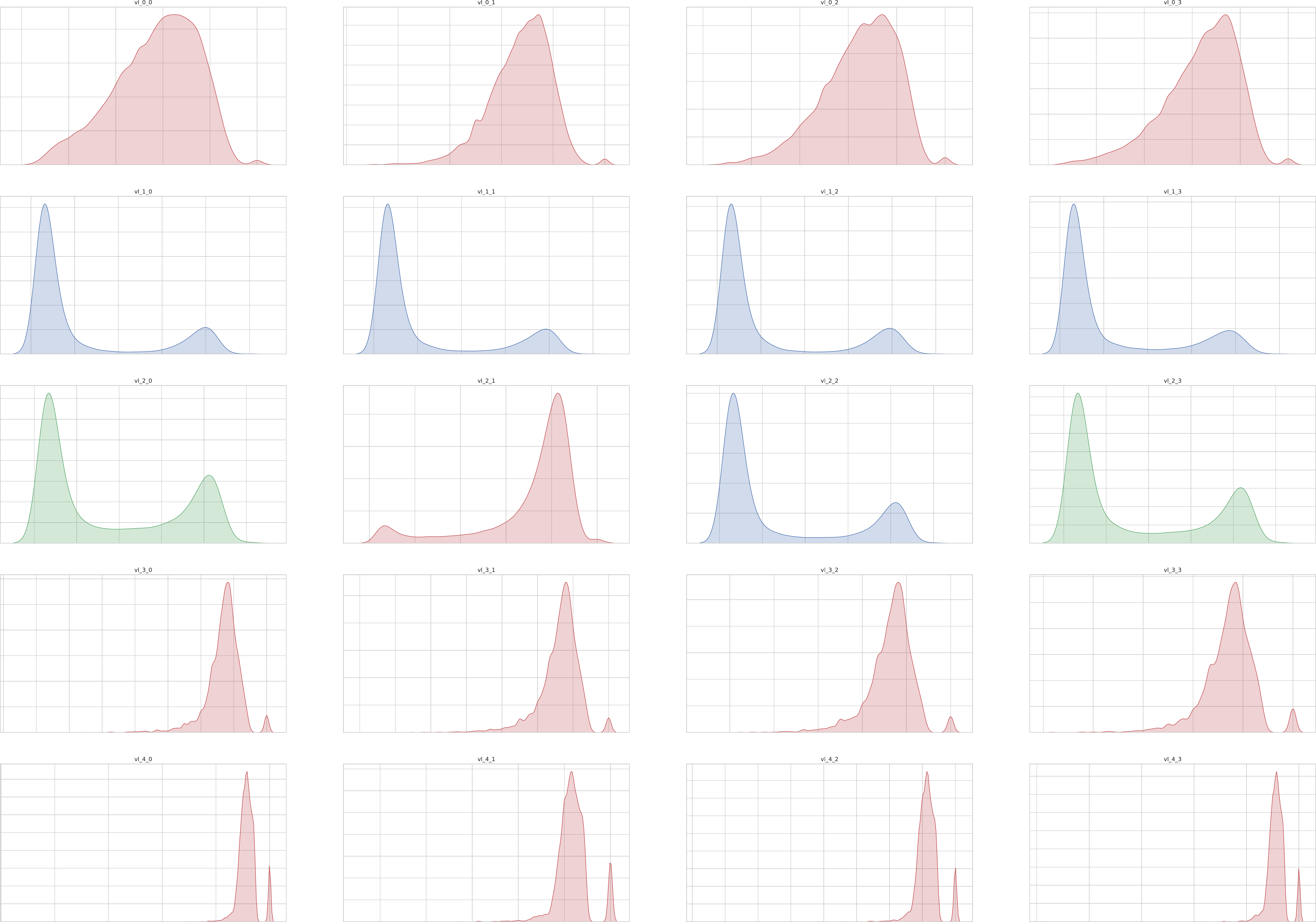}
    \end{minipage}
    \\
    \\
    (b) & \begin{minipage}{14cm}
        \includegraphics[width=\linewidth]{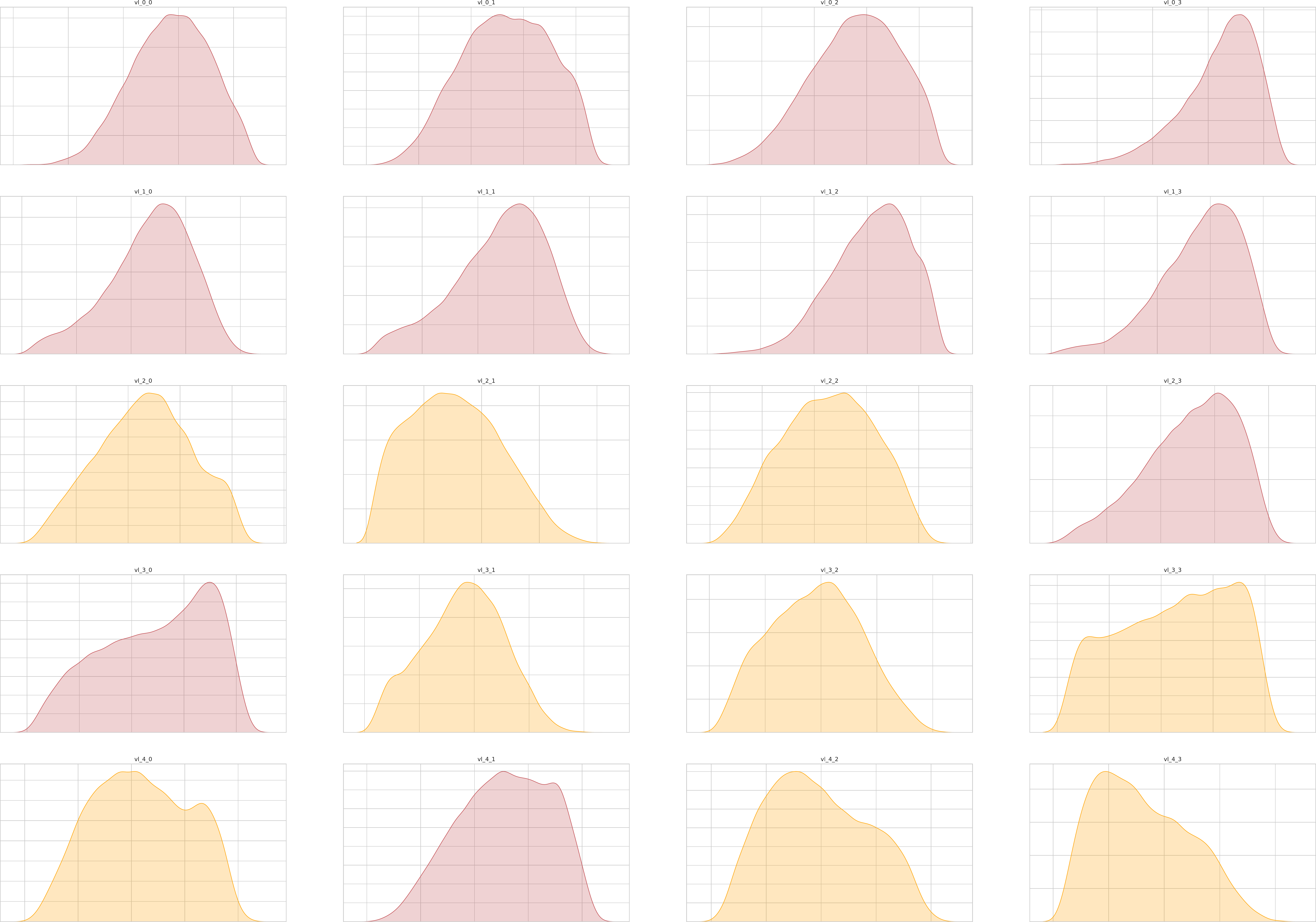}
    \end{minipage}
    \\
\end{tabular}
}
\vspace*{1mm}
\caption{\label{fig:vl_k_head_2}Comparison of k-distribution of VL-attention heads for two different models for the function \textit{choose color}: (a) oracle (4 first rows); (b) noisy visual input (4 last rows). Heads are colored according to their $k$-number median. As a recall, for each head we plot the distribution of the number $k$ of tokens required to reach $90\%$ of the attention energy (GQA-val). 
\cam{The x-axis represents in $\%$ the number of tokens $k$ relatively to the total number of token, it goes from $0\%$ to $100\%$.}}
\end{figure*}

\myparagraph{Attention distribution}
To give a better insight of the differences in attention modes between oracle and noisy visual models, we provide more visualizations in Fig~\ref{fig:vl_k_head_2}. These plots are measured on the $t_{\times}^{L\leftarrow V}(\cdot)$ attention heads for questions involving to choose a color. We recommend the reader to compare Fig~\ref{fig:vl_k_head_2} with \emph{Fig~4 in the main paper}, to better understand the influence of the \textit{choose color} function.
We observe that the oracle's attention heads are dependant on the functions involved in the question. In particular, the bi-morph heads become either \textit{dirac} or \textit{uniform} depending on the function. In contrast, the attention heads of the noisy visual model remain identical regardless of the function.

\begin{table}[] \centering
    {\footnotesize
    \begin{tabular}{lccc}
        \toprule
        Layer & & Notation & Abbrev.\\
        \midrule
        $9\times $Language only & L $\leftarrow$ L & $T_{-}^{L}(\cdot)$ & $lang,i,j$\\[2mm]
        $5\times $Vision only & V $\leftarrow$ V & $T_{-}^{V}(\cdot)$ & $vis,i,j$ \\[2mm]
        \multirow{2}{*}{$5\times $Cross-modal} & L $\leftarrow$ V & 
        \Big\{ $\begin{matrix}
        T_{\times}^{L\leftarrow V}(\cdot)\\
        T_{-}^{L}(\cdot)
        \end{matrix}$ & 
        $\begin{matrix}
        vl,i,j\\
        ll,i,j
        \end{matrix}$ \\[4mm]
        & V $\leftarrow$ L &
        \Big\{ $\begin{matrix}
        T_{\times}^{V\leftarrow L}(\cdot)\\
        T_{-}^{V}(\cdot)
        \end{matrix}$ & 
        $\begin{matrix}
        lv,i,j\\
        vv,i,j
        \end{matrix}$  \\
        \bottomrule
    \end{tabular}
    }
    \vspace*{1mm}
    \caption{\label{tab:arch} VL-Transformer architecture and notation. A schematic view of the transformer is given in Fig~\ref{fig:vl_transformer}.}
\end{table}

\section{Technical details}

\myparagraph{VL-Transformer architecture}
More information about the VL-Transformer architecture can be found in the schematic illustration drawn in Fig~\ref{fig:vl_transformer}. All notations and abbreviations are summarized in Table~\ref{tab:arch}. The chosen architecture is similar to LXMERT~\cite{tan2019lxmert}.

\myparagraph{Training details}
All models were trained with the Adam optimizer~\cite{kingma2014adam}, a learning rate of $10^{-4}$ with warm starting and learning rate decay.
Training was done on one P100 GPU. Two P100 GPUs were used for BERT/LXMERT~\cite{tan2019lxmert} pre-training. 
For the oracle, the batch size was equal to $256$. We train during 40 epochs and select the best epoch using accuracy on validation.
The oracle transfer follows exactly the same procedure, except when using LXMERT pretraining.

In that case,  BERT/LXMERT~\cite{tan2019lxmert} pretraining is performed during 20 epochs max with a batch size of $512$. All pretraining losses are added from the beginning, including the VQA one.
Note that LXMERT~\cite{tan2019lxmert} is originally pre-trained on a corpus gathering images and sentences from MSCOCO~\cite{lin2014microsoft} and VisualGenome~\cite{krishna2017visual}. As the GQA dataset is built upon VisualGenome, the original LXMERT pre-training dataset contains samples from the GQA validation split. Therefore, \emph{we removed these validation samples from the pre-training corpus}, in order to be able to validate on the GQA validation split.

After pre-trainning, we finetune either on GQA~\cite{hudson2019gqa} or VQAv2~\cite{goyal2017making}.
For GQA, we finetune during $4$ epochs, with a batch size of $32$ and a learning rate equal to $10^{-5}$.
For VQAv2, we finetune during $8$ epochs, with a batch size of $32$ and a learning rate equal to $10^{-5}$.

\end{document}